\documentclass[journal]{IEEEtran}
\pdfoutput=1
\usepackage{times}
\usepackage{epsfig}
\usepackage{graphicx}
\usepackage{amsmath}
\usepackage{amsthm}
\usepackage{amssymb}
\usepackage{bm}
\usepackage{amsfonts}
\usepackage{subfigure}
\usepackage{caption}
\usepackage{mathabx}
\usepackage{authblk}
\usepackage{xcolor}

\usepackage[utf8]{inputenc}
\ifdefined\DeclareUnicodeCharacter
  \DeclareUnicodeCharacter{0301}{\nobreakspace}
\fi

\usepackage{hyperref}
\hypersetup{colorlinks,allcolors=black}

\title{DefSLAM: Tracking and Mapping of Deforming Scenes from Monocular Sequences}

\author[1]{J. Lamarca}
\author[2]{S. Parashar}
\author[2]{A. Bartoli}
\author[1]{J.M.M. Montiel, \textit{Member}, \textit{IEEE}}

\affil[1]{\footnotesize Instituto de Investigación en Ingeniería de Aragón (I3A), Universidad de Zaragoza,Spain, \{jlamarca,josemari\}@unizar.es}
\affil[2]{\footnotesize Institut Pascal - UMR 6602 - CNRS/UCA/CHU, Clermont-Ferrand, France, \{adrien.bartoli,shaifali.parashar\}@gmail.com}

\begin{document}
\thispagestyle{empty}
{
\centering
\onecolumn
This paper has been accepted for publication in IEEE Transactions on Robotics.

\textcopyright 2020 IEEE. Personal use of this material is permitted. Permission from IEEE must be obtained for all other uses, in any current or future media, including reprinting/republishing this material for advertising or promotional purposes, creating new
collective works, for resale or redistribution to servers or lists, or reuse of any copyrighted component of this work in other
works.
}
\twocolumn[{%
\renewcommand\twocolumn[1][]{#1}%
\maketitle
\begin{center}
    \centering
    \includegraphics[width=\textwidth]{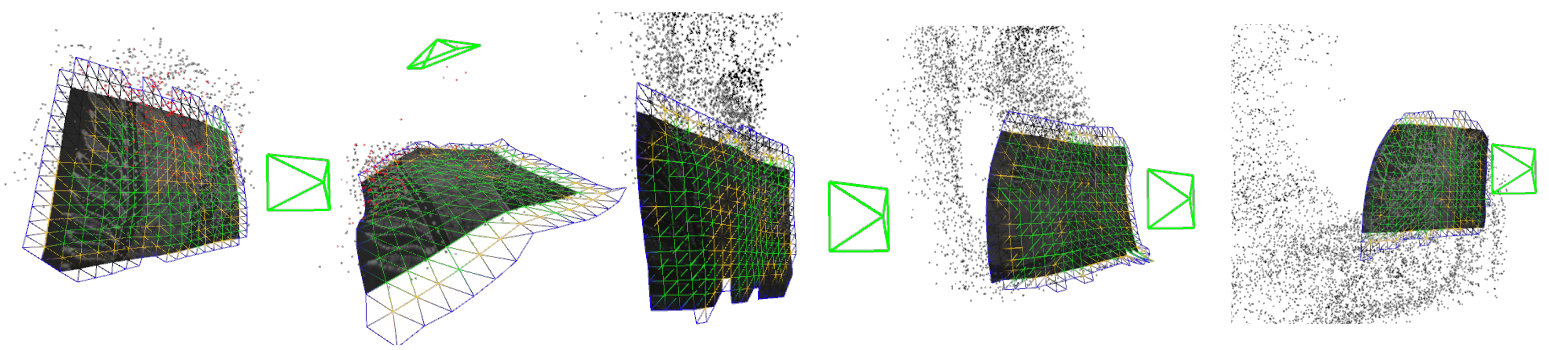}
    \captionof{figure}{Real-time reconstruction of a deforming scene with DefSLAM. The mandala kerchief deforms while the camera moves. DefSLAM locates the camera shown as a green frustum, while recovering the deformation of the kerchief using a template of the same. The estimated 3D deformable map is expanded when new regions are explored by reestimating new templates. The map is composed of sparse 3D points, in black, and a template as triangular mesh, viewed part in green.}
    \label{fig:overall_figure}
\end{center}
}]
\begin{figure*}[!h]
    \centering
    
\end{figure*}

\begin{abstract}
Monocular SLAM algorithms perform robustly when observing rigid scenes, however, they fail when the observed scene deforms, for example, in medical endoscopy applications. We present  \mbox{DefSLAM}, the first monocular SLAM capable of operating in deforming scenes in real-time. Our approach intertwines Shape-from-Template (SfT) and \mbox{Non-Rigid} \mbox{Structure-from-Motion} (NRSfM) techniques to deal with the exploratory sequences typical of SLAM. A deformation tracking thread recovers the pose of the camera and the deformation of the observed map, at frame rate, by means of SfT processing a template that models the scene shape-at-rest. A deformation mapping thread runs in parallel with the tracking to update the template, at keyframe rate, by means of an isometric NRSfM processing a batch of full perspective keyframes. In our experiments, DefSLAM processes close-up sequences of deforming scenes, both in a laboratory controlled experiment and in medical endoscopy sequences, producing accurate 3D models of the scene with respect to the moving camera.
\end{abstract}

\section{Introduction}
The goal of visual Simultaneous Localization and Mapping (SLAM) algorithms is to locate a visual sensor in an uncertain map which is being estimated simultaneously. The typical use case in SLAM includes exploratory trajectories where the camera images a scene without previous information of the structure observed. Using a monocular sensor, visual SLAM has to process several images rendering enough parallax to recover the map for the new scene region \textit{wrt.} the camera. Once the map is available, the camera can be localized \textit{wrt.} this map from just one image as long as the camera does not move to unexplored areas. The rigidity assumption constrains the problem significantly, and it is intensively exploited by state-of-the-art monocular SLAM systems \cite{engel2017direct,klein2007parallel,mur2015orb}.

However, the rigidity assumption rends invalid in applications where the deformation is predominant. To this end, we introduce \textbf{DefSLAM}, a calibrated monocular and deformable SLAM system which can perform in deforming, i.e. non rigid, environments. A relevant use case is medical endoscopy, where monocular visual SLAM is crucial a tool for augmented reality and autonomous medical robotics.

In the literature, non-rigid monocular scenes have been handled by Non-Rigid Structure-from-Motion (NRSfM)~\cite{chhatkuli2014non,chhatkuli2016inextensible,parashar2017isometric,taylor2010non,vicente2012soft} and Shape-from-Template (SfT)~\cite{chhatkuli2017stable,lamarca2018,ngo2016template,salzmann2011linear} methods. NRSfM methods are able to  recover the evolution of the 3D scenes non-rigid deformations from a set of monocular images, after a computationally demanding batch processing of the images. In contrast, SfT recovers the 3D deformation from a single image, at a low computational cost but needs a template. The template is a 3D textured model describing the shape at rest of the scene. DefSLAM framework combines the advantages of the two classes of non-rigid monocular methods. We propose a parallel algorithm composed of a
\emph{deformation tracking} thread as the front-end running SfT at frame rate, and a \emph{deformation mapping} thread as the back-end running NRSfM to compute the SfT template at a slower keyframe rate. 

Fig.\, \ref{fig:overall_figure} shows DefSLAM processing a sequence processed where the camera is being located \textit{wrt.} a deforming kerchief being mapped simultaneously using images from a monocular sensor from partial observations of different regions of the kerchief. The \emph{deformation tracking} thread recovers the camera pose and the deformation of the map at frame rate. It uses a template for the viewed part of the map to recover the map points deformation by minimizing a combination of reprojection error and deformation energy for each frame. The \emph{deformation mapping} thread initializes and refines map estimates, and extends the map when new regions are visited. It processes just a selection of frames -- keyframes -- imaging the same region to define the shape-at-rest of the template used by the \emph{deformation tracking} thread to process the subsequent frames.

We validated our \mbox{DefSLAM} algorithm in monocular sequences that include exploratory trajectories observing deforming scenes. We evaluate DefSLAM on new waving mandala kerchief dataset which we created and an in-vivo medical endoscopy Hamlyn dataset \cite{mountney2010three}. To make some comparison we have resorted to systems with a different configuration than ours. We compare our results with the state-of-the-art rigid monocular \mbox{ORBSLAM}\,\cite{mur2015orb} to display the DefSLAM unique capability to SLAM deforming scenes. We also compared with \mbox{MISSLAM}\,\cite{song2018mis}, the closest in the literature offering SLAM accuracy results in medical deformable scenes, despite it is stereo in contrast to our monocular system. These experiments validate the unprecedented ability of \mbox{DefSLAM} to accurately code the structure of the scene in rigid and deformable scenarios, including medical cases.

\section{Related Work} 
\label{Related} 
\subsection{SLAM}
\textbf{Deformable visual SLAM.} The deformable SLAM methods in the literature rely on sensors providing depth information, \textit{i.e.} RGB-D or stereo sensors. DynamicFusion \cite{newcombe2015dynamicfusion} is a seminal work in deformable VSLAM with an RGB-D camera. It fuses the frame-by-frame depth information into a canonical shape, i.e. a shape at rest, that incrementally maps the entire scene after an exploratory trajectory of partial observations. This canonical shape is deformed to the current keyframe with the as-rigid-as-possible deformation model \cite{sorkine2007rigid}. In \cite{innmann2016volumedeform}, the quality of the deformation is improved by including the photometric error in the optimization. In \cite{gao18surfelwarp}, the volumetric representation is substituted by surfels to improve the efficiency of the algorithm. These methods recover the whole canonical shape deformation which is usually small. This technique is not scalable to bigger shapes like exploratory scenes in endoscopy. \cite{song2017dynamic} proposes to use an embedded deformation model \cite{sumner2007embedded} instead of as-rigid-as-possible because it better preserves the local details under the deformation. In \cite{song2018mis}, the system is enhanced with the tracking of a rigid system ORBSLAM \cite{mur2015orb} to achieve better tracks and more robust deformable SLAM for medical endoscopy exploration. In any case, all these algorithms optimize the whole map each time and thus scale poorly with the size of the map.  We aim similar SLAM capabilities in deformable scenes, but in the challenging monocular case. In addition, our approach only optimizes the observed map zone achieving good scalability \textit{wrt.} the size of the map, being able to be run on the CPU.

\textbf{Rigid visual SLAM.} Monocular rigid VSLAM is a mature field. The current state-of-the-art monocular rigid VSLAM methods such as \cite{engel2017direct,mur2015orb} provide accurate, robust and fast results in robotic scenes. Some works have attempted to apply rigid methods in in-vivo medical quasi-rigid scenes. \cite{grasa2014visual} proposes an EKF-SLAM algorithm, and \cite{mahmoud2018live} gets dense maps based on \cite{mur2015orb}. \cite{marmol2019dense} uses a rigid SLAM system to locate the camera in arthroscopic images. All of these methods assume that the deformation is negligible and hence that a purely rigid SLAM system is able to survive just by excluding from the map any deformed scene region. We aim to achieve a similar performance, but in scenarios where deformation is predominant, more specifically: real-time operation and capability to handle sequences of close-ups corresponding to exploratory trajectories.

\subsection{Non-Rigid Monocular Techniques}
The methods in the literature which aim to recover the structure of a non-rigid scene from monocular sequences are SfT and NRSfM.  

\textbf{Shape-from-Template.} SfT methods recover the deformed shape of an object from a monocular image and the object's textured 3D shape at rest. This textured shape-at-rest of the object is the so-called \textbf{template}. These methods associate a deformation model with this template to recover the deformed shape. The main difference between these methods is the definition of the deformation model. We distinguish between analytic and energy-based methods. Among the analytic solutions, we focus on the isometric deformation which assumes that the geodesic distance between points in the surface is preserved. Isometry for SfT has proven to be well-posed and it quickly evolved to stable and real-time solutions  \cite{bartoli2015shape,chhatkuli2017stable,collins2010locally}. Energy-based methods \cite{agudo2014good,lamarca2018,ngo2016template,salzmann2011linear} jointly minimize the shape energy  \textit{wrt.} the shape-at-rest and the reprojection error for the image correspondences. These optimization methods are well suited to implement sequential data association with robust kernels to deal with outliers.

\textbf{Orthographic Non-Rigid Structure-from-Motion.} The earliest non-rigid monocular techniques are NRSfM. These methods were formulated using statistical models, first proposed in \cite{bregler2000recovering}. This work gave rise to a family of methods \cite{dai2014simple,moreno2011probabilistic,paladini2009factorization} which used a low dimensional basis model to obtain the configuration of the 3D points from the images of a sequence. They exploited spatial reguralizers \cite{dai2014simple,garg2013dense}, temporal regularizers \cite{akhter2011trajectory} and spatio-temporal regularizers \cite{agudo2015simultaneous,gotardo2011kernel,gotardo2011non}. These methods may handle small surface deformations or articulated objects, but they usually fail with very large deformations. They use an orthographic camera model which is an approximation only valid when the scene is distant from the camera, this is a strong assumption invalid in many applications.

\textbf{Perspective Non-Rigid Structure-from-Motion.} Real use cases need the more accurate perspective camera model. It is able to model the close-up sequences typical in SLAM, especially in medical endoscopy. The isometry assumption, first proposed in SfT methods, has also produced excellent results in NRSfM \cite{chhatkuli2014non,chhatkuli2016inextensible,parashar2017isometric,taylor2010non,vicente2012soft}. It brought not only improvements in terms of accuracy, but also the ability to handle perspective cameras.  \cite{parashar2017isometric} is a local method, able to handle naturally occlusions and missing data also usual in many applications. 

\textbf{Our approach.} We propose the first visual SLAM system capable of working with deforming monocular sequences. We propose a \emph{deformation tracking} thread based on \cite{lamarca2018}, which uses a pre-computed template to recover the camera pose and the deformation of the scene. We also propose a \emph{deformation mapping} thread which extends the map and estimates the shape-at-rest of the template in new explored zones by means of the isometric NRSfM proposed by \cite{parashar2017isometric}. Our contribution is a new iterative scheme for the optimization in \cite{parashar2017isometric} that allows to calculate and refine the solutions incrementally at keyframe rate. Both for the deformation mapping and tracking, we only optimize the part of the template observed having a runtime independent of the size of the map in exploratory sequences.

We also propose a sequential active matching that exploits the already available SLAM map to boost the data association performance. Our final contribution is to integrate in the deformation mapping an alignment between surfaces to build a global map, extending alignment as proposed in \cite{newcombe2015dynamicfusion,gao18surfelwarp,song2018mis} to the monocular case.

The proposed deformation tracking and mapping algorithms can run in parallel, in a similar way to the state-of-the-art rigid SLAM methods \cite{engel2017direct,klein2007parallel,mur2015orb} to achieve real-time performances. 
\section{DefSLAM System Overview}
\label{sec:pipeline}
DefSLAM recovers the structure of the scene, its deformation and the camera pose. It is composed of three main components: 

\begin{itemize}
\item \textbf{The map}. The map represents the structure of the scene reconstructed by DefSLAM as a set of 3D map points. The map is deformable and the position of the map points evolves along the sequence. Each map point $j$ is represented by its position ${\mathbf{X}_j^t}$ for each processed frame $t$.

We save some selected frames in the map called keyframes. We refer to the keyframes in which a map point is initialized as anchor keyframes. After each new keyframe processing, one of the anchor keyframes is selected as the reference keyframe. The reference keyframe defines the template used by the deformation tracking to process the new incoming frames.

\item \textbf{The deformation tracking thread}. This thread is the front-end of the system and runs at frame rate. It uses SfT to estimate the position of the map points ${\mathbf{X}_j^t}$ and the camera pose $\mathbf{T}_{tw}$ for each frame $t$. We embed the map points into the template $\mathcal{T}_k$ to compute their position. The shape-at-rest of the template $\mathcal{T}_k$ is the surface $\mathcal{S}_k$ observed in the reference keyframe $k$.

\item \textbf{The deformation mapping thread}. This thread is the back-end of the system and runs at keyframe rate. It uses NRSfM to estimate the surface $\mathcal{S}_k$ observed in the keyframe $k$. 
\end{itemize}

\textbf{Notation} We use calligraphic letters for sets of geometrical entities in the deforming scene, \textit{e.g.} $\mathcal{X}$ for the set of all map points. Bold letters represent matrices and vectors. Scalars are represented in italics. The indexes $t$ represent the frames and $\mathbf{T}_{tw}$ the pose of the frame at instant t. Superindexes represent the temporal instant of the estimation. The index $j$ represents the map points, $n$ the nodes and $e$ the edges of the mesh describing the template surface.

\section{Deformation Tracking}
\label{sec:DefTrack}
\textbf{Deformation tracking} recovers the camera pose $\mathbf{T}_{tw}$ and the shape of the template $\mathcal{T}_{k}^t$ in the frame $t$ by jointly minimizing reprojection error and deformation energy.  $\mathcal{T}_{k}$ is the surface reconstructed in the reference keyframe $k$. The tracking algorithm is composed of three stages: data association, camera pose estimation and template deformation and new keyframe selection. Next, it is detailed the template structure, the camera model and the three steps of the algorithm.

\subsection{Template}
The template is a surface parametrized with a 3D triangular mesh. It is composed of a set of planar triangular facets $\mathcal{F}$, defined by a set of nodes $\mathcal{V}$, and connected by a set of edges $\mathcal{E}$. The deformation of the map at frame $t$ is defined through the pose of the nodes of the template $\mathcal{T}_{k}^t$. The facet $f \in \mathcal{F}$ at frame $t$ is defined by the pose of its three nodes $V_{f_j}^{t} = \{V_{f,h}^{t}\},\;\; h=\{1,2,3\}$. The map points observed in the keyframe $k$ are embedded in the facets of the mesh. The position of a map point $\textbf{X}_j^t \in \mathcal{X}$ in frame $t$ is defined with its barycentric coordinates, $\bm{b}_{j}=\left[b_{j,1}, b_{j,2} ,b_{j,3}\right]^\top$, \textit{wrt}. the position of the nodes of the face $f_j$ :

\begin{equation} \label{eq:Embeddingfunction}
\bm{X}_{j}^{t} =  \sum_{h=1}^3 b_{j,h} \bm{V}_{f_j,h}^{t} \,\, \mbox{s.t.} \,\, b_{j,1}+b_{j,2}+b_{j,3} = 1.
\end{equation}

\subsection{Camera Model}
We use the calibrated pinhole model. The projection of the 3D point $j$, $\bm{X}_{j}^{t} \in \mathcal{X}_k^t$ in the frame $t$ by a camera located at $\bm{T}_{tw}$ is modelled by the projection function \mbox{$\pi:[\mbox{SE}\left(3\right),\mathbb{R}^3] \to \mathbb{R}^{2}$}:

\begin{eqnarray}\label{eq:ProjectionFunction}
\pi\left(\bm{T}_{tw},\bm{X}_{j}^{t}\right) &=& \left[\begin{array}{c}f_{x}\frac{X_{j}^{t}}{Z_{j}^{t}}+C_{x} \label{eq:perspective_SLAM}\\
            f_{y}\frac{Y_{j}^{t}}{Z_{j}^{t}}+C_{y}\end{array}\right],\\ \text{where } &&
                     \left[\begin{array}{ccc} X_{j}^{t}&Y_{j}^{t}&Z_{j}^{t}\end{array}\right]^\top =\bm{R}_{tw} \bm{X}_{j}^{t} + \bm{t}_{tw}. \nonumber
\end{eqnarray}$\bm{R}_{tw} \in SO(3)$ and $\bm{t}_{tw} \in \mathbb{R}^3$ are respectively the rotation and the translation of the transformation $\bm{T}_{tw}$. $\left\{f_{x},f_{y},C_{x},C_{y}\right\}$ are the focal lengths and the principal points from the camera calibration. The set of observation in the image $\mathcal{I}^t$ are the keypoints $\textit{x}^t$ matched with a map point of $\mathcal{X}^t$. The map point $\textbf{X}_j^t$ is projected in the normalized retina as $(\hat{x}_j^t,\hat{y}^t_j)$ where $\hat{x}^t_j=\frac{x_j^t-C_x}{f_x}$, $\hat{y}^t_j=\frac{y_j^t-C_y}{f_y}$ and  $ \begin{pmatrix}
x_j^t &y_j^t
\end{pmatrix}^\top =\pi\left(\mathbf{T}_{tw},\mathbf{X}_{j}^{t}\right) $.

\subsection{Camera Pose and Template Deformation}
\begin{figure}
    \centering
    \includegraphics[width = \columnwidth,height=0.7\columnwidth]{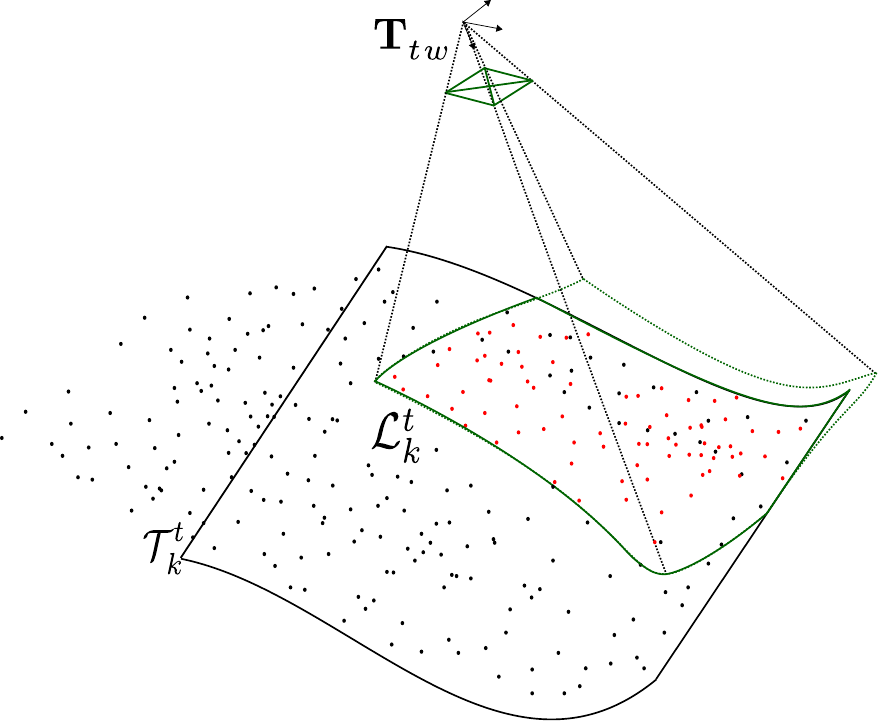}
    \caption{Deformation tracking: estimating camera pose and deformation of the viewed map. $\mathcal{T}_{k}^t$ is the map shape in the frame $t$, $\mathcal{L}_{k}^t$ is the local map shape in the frame $t$ and $\textbf{T}_{cw}^t$ the camera pose. Black points belong to the global map. Some of them are embedded in the template. Current matched points in red.}
    \label{fig:deformationtracking}
\end{figure}
In SLAM sequences, the camera usually images a zone smaller than the template. For efficiency and scalability, we only optimize the observed zone of the template and its closest vicinity. We refer to this part of the template as the local zone $\mathcal{L}_{k}^t \subseteq \mathcal{T}_{k}^t$. Figure\,\ref{fig:deformationtracking} shows all the components of the deformable tracking: the template $\mathcal{T}_{k}^t$, the local zone ${\mathcal{L}_{k}^t}$ and the camera pose $\mathbf{T}_{tw}$. 

To estimate the deformed ${\mathcal{L}_{k}^t}$ and $\mathbf{T}_{tw}$, we jointly minimize the reprojection error $\varphi_{d}(\mathcal{I}^t,\mathbf{T}_{cw},\mathcal{L}_{k}^t)$ in the image $I^t$ and the deformation energy $\varphi_{e}(\mathcal{L}_{k}^t, \mathcal{T}_k)$ of the template $\mathcal{T}_k$:

\begin{eqnarray}
\label{eq:SfTLamarca}
 &\underset{\mathcal{L}_{k}^t,\mathbf{T}_{tw}}{\arg\min}  &  \varphi_{d}(\mathcal{I}^t,\mathbf{T}_{tw},\mathcal{L}_{k}^t) + \varphi_{e}(\mathcal{L}_{k}^t,\mathcal{T}_k).
\end{eqnarray}
We solve (\ref{eq:SfTLamarca}) using the Levenberg-Marquardt optimization method. The initial guess for $(\mathcal{L}_{k}^t,\mathbf{T}_{tw})$, is the solution of the previous frame, $(\mathcal{L}_{k}^{t-1},\mathbf{T}_{t-1 w})$. We fix the pose boundary nodes of ${\mathcal{L}_{k}^t}$ during the optimization to constraint the gauge freedoms of the camera pose $\mathbf{T}_{tw}$, 

The reprojection error $\varphi_{d}(\mathcal{I}^t,\mathbf{T}_{tw},\mathcal{L}_{k}^t)$ for the set of keypoints $\textit{x}^t$  in image $\mathcal{I}^t$ is defined as:

\begin{equation}
\label{fig:rep_error}
\varphi_{d}(\mathcal{I}^t,\mathbf{T}_{tw},\mathcal{L}_{k}^t) = \sum_{j\in \textbf{\textit{x}}^t} \rho \left (\left\|\pi(\mathbf{X}_{j}^{t},\textbf{T}_{tw}) - \mathbf{x}_j^t\right\|\right).
\end{equation}
The reprojection error is robust against outliers as it is weighted with a Huber robust kernel $\rho(.)$.

We define a deformation energy $\varphi_{e}(\mathcal{L}_{k}^t,\mathcal{T}_{k})$ \textit{wrt}. $\mathcal{T}_{k}$ as a combination of a stretching energy $\varphi_{s}(\mathcal{L}_{k}^t,\mathcal{T}_{k})$, a bending energy $\varphi_{b}(\mathcal{L}_{k}^t,\mathcal{T}_{k})$ and a reference regularizer $\varphi_{r}(\mathcal{L}_{k}^t,\mathcal{T}_k)$:

\begin{equation}
\label{eq:deformation_Energy}
\begin{split}
\varphi_{e}(\mathcal{L}_{k}^t,\mathcal{T}_{k}) = & \,\, \lambda_s \varphi_{s}(\mathcal{L}_{k}^t,\mathcal{T}_{k}) +
\lambda_b \varphi_{b}(\mathcal{L}_{k}^t,\mathcal{T}_{k})\\
&
+\lambda_{r}\varphi_{r}(\mathcal{L}_{k}^t,\mathcal{T}_{k}).
\end{split}
\end{equation}
We use $\lambda_s$, $\lambda_b$ and $\lambda_{r}$ to weight the influence of each term. 

The stretching energy $\varphi_{s}(\mathcal{L}_{k}^t,\mathcal{T}_{k})$ measures the difference in the length $l_e^t$ of each edge $e$ in the local zone $\mathcal{L}_{k}^t$  in the frame $t$ with respect to its length $l_e^k$ in the shape-at-rest of $\mathcal{T}_{k}$:

\begin{equation}
    \varphi_{s}(\mathcal{L}_{t}^k,\mathcal{T}_{k}) = \sum_{e \in \mathcal{L}_k^t}\left(\frac{l_e^t-l_e^k}{l_e^k}\right)^2.
    \label{eq:stretching_reg}
\end{equation}

The bending energy $\varphi_{b}(\mathcal{L}_{k}^t,\mathcal{T}_{k})$ measures the changes in mean curvature ${\delta}_{n}^{t}$ in each node $n$ \textit{wrt.} the estimated ${\delta}_{n}^{k}$ in the shape-at-rest of $\mathcal{T}_{k}$. We estimate the mean curvature through the discrete Laplacian operator \cite{floater2003mean}. We make the bending term dimensionless by dividing it by the mean distance $l_e^k$ of the edges connected with the node $\mathcal{E}_n^k$:

\begin{equation}
    \varphi_{b}(\mathcal{L}_{k}^t,\mathcal{T}_{k}) = \sum_{n \in \mathcal{L}_{k}^{t}} \sum_{e \in \mathcal{E}_n^k} \left(\frac{\delta_{n}^{t}-\delta_{n}^{k}}{l_{e}^{k}} \right)^2.
    \label{eq:bending_reg}
\end{equation}

Optimization considering the terms $\varphi_{d}(\cdot)$, $\varphi_{b}(\cdot)$ and $\varphi_{s}(\cdot)$ allows to recover the relative pose of the camera with respect to the template, but the absolute camera pose is not observable. Thanks to the fixation of the ${\mathcal{L}_{k}^t}$ boundary nodes pose, the absolute camera pose becomes observable. However, the camera pose sometimes is only weakly observable depending on the boundary nodes geometrical distribution and cardinality. If the template is completely observed by the camera, then there are no boundary points to be fixed and the camera pose becomes fully non-observable.

We add another regularizer, $\varphi_{r}(\mathcal{L}_{k}^t,\mathcal{T}_{k})$, that we call  reference regularizer to keep the template as close as possible to its initial position in its reference keyframe, to alleviate the camera pose weak observability. It is given by:
\begin{equation}
\varphi_{r} (\mathcal{L}_{k}^t,\mathcal{T}_{k}) = \sum_{n\in \mathcal{L}_{k}^{t}} \left\|\mathbf{V}_{n}^{t} -\mathbf{V}_{n}^{k}\right\|.
\label{eq:reference_reg}
\end{equation} 

Optimization (\ref{eq:SfTLamarca}) also needs the derivatives of the regularizers (\ref{eq:stretching_reg})-(\ref{eq:reference_reg}), they are detailed in Appendix  \ref{FirstAppendix}.

\subsection{Data Association}
\label{sec:datassoTrack}
To match the keypoints in the current frame with the map points, we apply an active matching strategy as proposed in \cite{davison2003real}. First, the ORB keypoints are detected in the current frame. Next, the camera pose is predicted with a camera motion model as a function of the past camera poses. Then, we use the last estimated shape of template and the barycentric coordinates to predict where the map points will be imaged. Around the map point prediction, we define a search region. We match the map point with the keypoint with the most similar ORB descriptor inside its search region. The similarity is estimated as the Hamming distance between the ORB descriptors, the match is accepted only if it is below a distance threshold. The ORB descriptor of the map point is taken from the keypoint of the keyframe where it was initialized.

\subsection{New Keyframe Selection}
We select a new keyframe as soon as the mapping thread finishes its processing. If the new keyframe covers a new map region, it becomes an anchor keyframe and the reference keyframe and a new template is created. Otherwise, the new keyframe is a regular keyframe, and its most covisible anchor keyframe is selected as the reference keyframe, and its template is refined.

\section{Deformation Mapping}
\label{sec:DefMap}
\begin{figure}[t]
    \centering
    \includegraphics[width =0.49\textwidth,height =0.7\columnwidth]{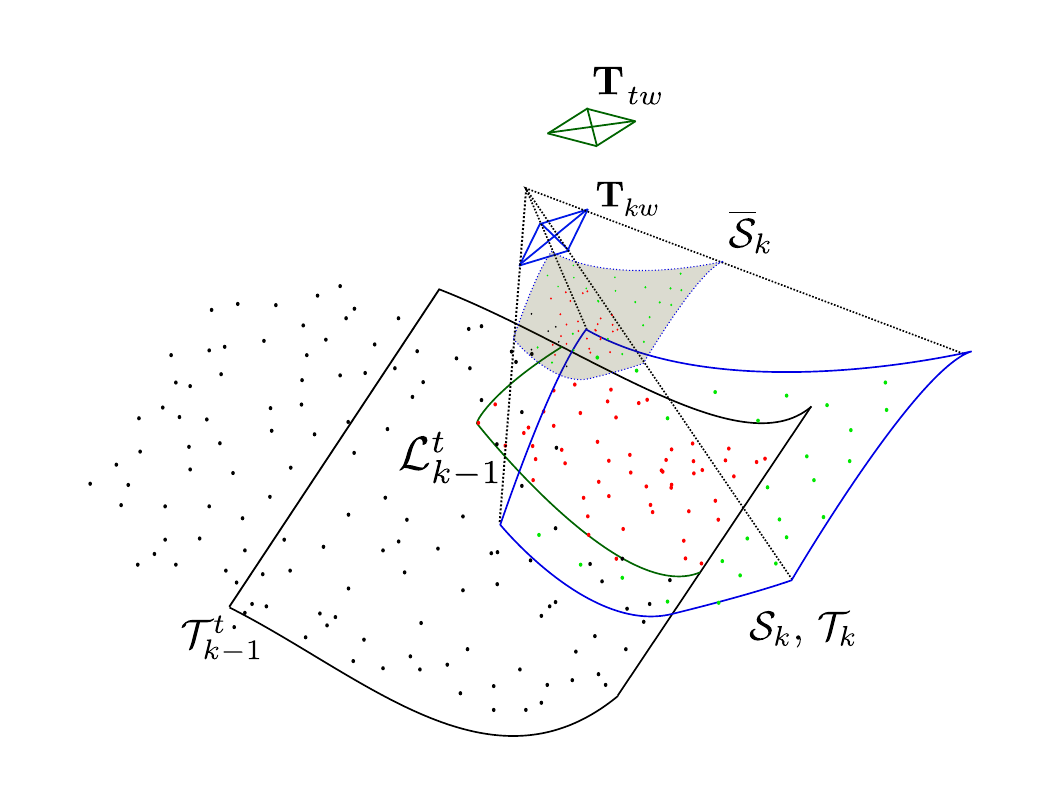}
    \caption{Extension of the map in the deformable mapping.
Local area $\mathcal{L}_{k-1}^t$ in green. Matched points in red. In blue, the up-to-scale surface estimated by NRSfM, $\overline{\mathcal{S}}_{k}$ (dotted line), and template $\mathcal{T}_{k}$ computed from the scaled surface $\mathcal{S}_{k}$ of the reference keyframe $k$.}
    \label{fig:MapPlusTrack}
\end{figure}

\textbf{Deformation mapping} recovers the observed map as a surface $S_{k}$ for the reference keyframe $k$. This surface contains the map points observed in the keyframe during the tracking. With the new keyframe we refine the map points and create new ones. $S_{k}$  defines the shape-at-rest of the template $\mathcal{T}_{k}$ for the deformation tracking for the next frames, as shown in Figure\,\ref{fig:MapPlusTrack}.

Deformation mapping is performed as follows: first, we compute the warps $\eta_{kk^*}$ between the anchor keyframes $k$ and the new keyframe $k^*$. At this stage, the considered anchor keyframes are those where one of the currently observed map points were intialized. Second, we estimate an up-to-scale surface $\overline{\mathcal{S}}_{k}$ by processing the covisible keyframes with the new keyframe by means of NRSfM. Third, we align $\overline{\mathcal{S}}_{k}$ with the previous map to recover the scale and the scaled surface $\mathcal{S}_k$. Finally, with this new surface, we create the new template by computing a triangular mesh and embedding the map points in its facets.

\subsection{NRSfM}
\label{sec:Iso}
In isometric NRSfM, the surface deformation is modelled locally for each point under the assumption of isometry and infinitesimal planarity. Assuming infinitesimal planarity, any surface is approximated as a plane at an infinitesimal level, while maintaining its curvature at the global level. Isometric NRSfM can handle both rigid and non-rigid scenes. Since we use a local method, it can handle missing data and occlusions inherently. We build on the isometric NRSfM proposed in \cite{parashar2017isometric}. For the sake of completeness, we summarize the formulation.
\begin{figure}
    \centering
\includegraphics[width = \columnwidth,height = 0.5\columnwidth]{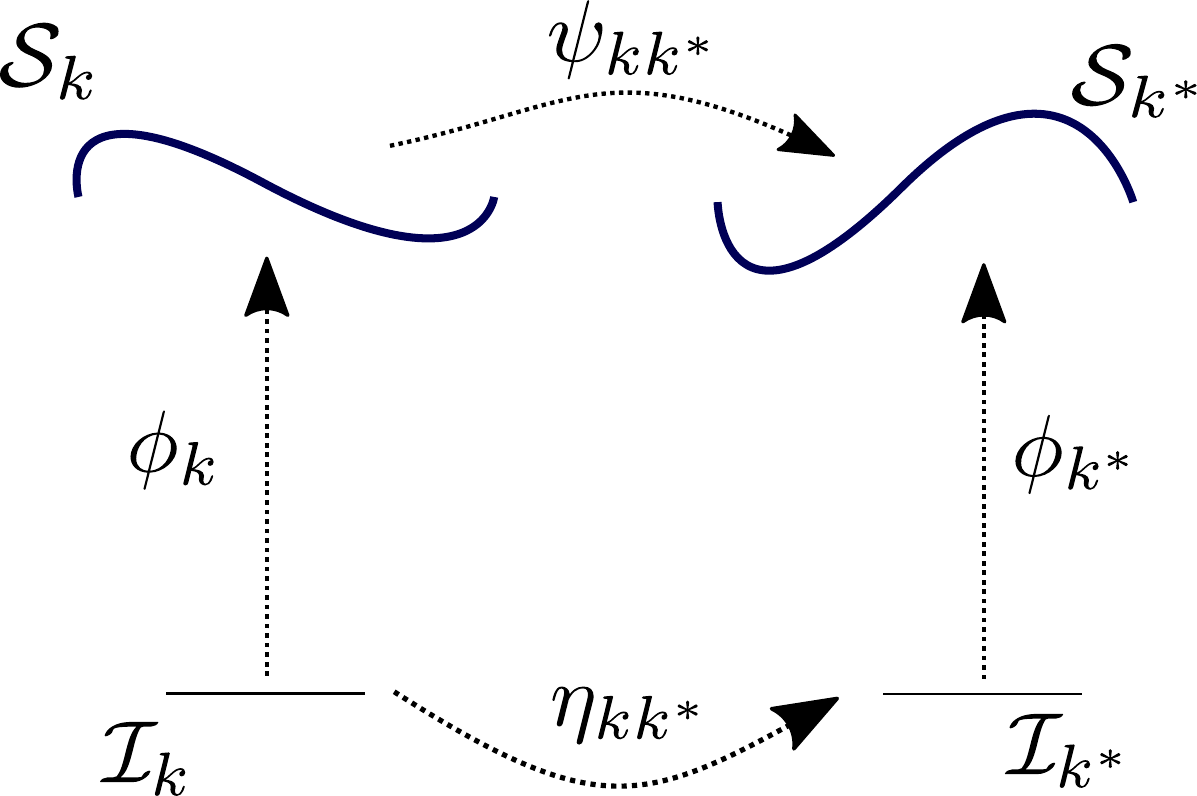}    
\caption{Relation between an anchor keyframe k and one of its covisibles ${k^*}$.  $\phi_{k}$ and $\phi_{k^*}$ are embeddings of the two   keyframe surfaces $k$ and $k^*$. $\eta_{k k^*}$ is the warp between $k$ and $k^*$. $\psi_{k k^*}$ is the deformation field between the surfaces $S_k$ and $S_{k^*}$}
\label{fig:IsoSystem}
\end{figure}

$\phi_k$ is the embedding of the scene surface $\mathcal{S}_k$, it is parametrized using the retina normalized coordinates of the image $\mathcal{I}_k$:
\begin{eqnarray}
\phi_k:\mathbb{R}^2 &\mapsto& \mathbb{R}^3 \nonumber
\\
\phi_k\left(\hat{x},\hat{y}\right) &=&
\begin{bmatrix}
\frac{\hat{x}}{\beta(\hat{x},\hat{y})} & \frac{\hat{y}}{\beta(\hat{x},\hat{y})} & \frac{1}{\beta(\hat{x},\hat{y})}
\end{bmatrix}^\top ,
\end{eqnarray}where $\beta_k(\hat{x},\hat{y})$ is the inverse depth of each point.
The normal $\vec{\mathbf{n}}_j(\hat{x},\hat{y})$ of the surface expressed \textit{wrt}. this parametrization is given as:
\begin{eqnarray}
\vec{\mathbf{n}}_j(\hat{x},\hat{y}) \propto \begin{pmatrix}
K_{\hat{x}}\\
K_{\hat{y}} \\
1- \hat{x} K_{\hat{x}}  - \hat{y} K_{\hat{y}}
\end{pmatrix},
\end{eqnarray}
where $K_{\hat{x}} = \frac{\beta_k(\hat{x},\hat{y})_{\hat{x}}}{\beta_k(\hat{x},\hat{y})}$ and $K_{\hat{y}}$ $= \frac{\beta_k(\hat{x},\hat{y})_{\hat{y}}}{\beta_k(\hat{x},\hat{y})}$, the subindexes $\hat{x}$ and $\hat{y}$ denote the partial derivatives.

NRSfM exploits the relationship between the metric tensor, $g_k(\hat{x},\hat{y})$, and the Christoffel symbols, $\Gamma_k^{\hat{x}}(\hat{x},\hat{y})$ and $\Gamma_k^{\hat{y}}(\hat{x},\hat{y})$, of the surface of the keyframe $\mathcal{S}_k$ and those of its covisible keyframes $\mathcal{S}_{k^*}$. Assuming infinitesimal planarity and isometry, $\Gamma_k^{\hat{x}}(\hat{x},\hat{y})$ and $\Gamma_k^{\hat{y}}(\hat{x},\hat{y})$ only depend on $K_{\hat{x}}$ and $K_{\hat{y}}$ for each point in every keyframe image. The warp $\eta_{k k^*}$ between the keyframes $k$ and $k^*$ represents the transformation from the image $\mathcal{I}_k$ to the image $\mathcal{I}_{k^*}$. Figure\, \ref{fig:IsoSystem} shows the different elements of the two view relation, the warp $\eta_{k k^*}$, the surface embeddings for each keyframe $\phi_k$ and $\phi_{k^*}$, and the isometric deformation $\psi_{k k^*}$ between the surfaces $\mathcal{S}_k$ and $\mathcal{S}_{k^*}$. Due to the infinitesimal planarity and isometry assumptions, the metric tensor and the Christoffel symbols in two different surfaces $k$ and $k^*$ are related through the warp between these keyframes $\eta_{k k^*}$ as:
\begin{eqnarray}
\label{eq:metric_rel}
  g_{k}(\hat{x},\hat{y}) = J_{\eta_{k k^*}}^\top g_{k^*}(\hat{x}^*,\hat{y}^*) J_{\eta_{k k^*}} 
\\
\Gamma_k^q(\hat{x},\hat{y})= \sum_h \frac{\partial \hat{x}_h}{\partial \hat{x}_h^*}(J_{\eta_{k k^*}}^\top \Gamma_k^h(\hat{x}^*,\hat{y}^*) J_{\eta_{k k^*}} +  H_{\eta_{k k^*}}^h),
\label{eq:Chris_rel}
\end{eqnarray}
where $J_{\eta_{k k^*}}$ and $H_{\eta_{k k^*}}^q$ are the Jacobian and the Hessian for the variable $q=\{\hat{x},\hat{y}\}$ of the warp $\eta_{k k^*}$ respectively. 
Eqs.\,(\ref{eq:metric_rel}) and (\ref{eq:Chris_rel}) can be transformed in two cubic polynomial equations
 $P(K_{\hat{x}}^k,K_{\hat{y}}^k)$ and $Q(K_{\hat{x}}^k,K_{\hat{y}}^k)$ for each point correspondence:
\begin{eqnarray}
  P(K_{\hat{x}}^{k},K_{\hat{y}}^{k}) &=& \sum_{u,v \in [0,3]} p_{uv} (K_{\hat{x}}^{k})^u (K_{\hat{y}}^{k})^v = 0 \label{eq:P}\\
  Q(K_{\hat{x}}^{k},K_{\hat{y}}^{k}) &=& \sum_{u,v \in [0,3]} q_{uv} (K_{\hat{x}}^{k})^u (K_{\hat{y}}^{k})^v = 0, \label{eq:Q}
\end{eqnarray}
where the coefficients $p_{uv}$ and $q_{uv}$ depend only on the normalized coordinates of the points and the derivatives of first and second-order derivatives of the warp $\eta_{k k^*}$. We refer to \cite{parashar2017isometric} for further details in the coefficients $p_{uv}$ and $q_{uv}$. 

\subsection{Incremental Surface Normals Refinement}
\label{sec:incremOpt}
If a point is matched in two or more keyframes,
we can calculate its normal in its anchor keyframe $k$, defined by $K^{k}_{\hat{x}}$ and $K^{k}_{\hat{y}}$, by means of non-linear optimization:
\begin{equation}
    \underset{K_{\hat{x}}^k,K_{\hat{y}}^k}{\arg \min}\,\,    \left(P\left(K_{\hat{x}}^{k},K_{\hat{y}}^{k}\right)\right)^2+ \left(Q  \left(K_{\hat{x}}^{k},K_{\hat{y}}^{k}\right)\right)^2.
    \label{eq:noptimization}
\end{equation}

\begin{figure}[t]
    \centering
\includegraphics[width = 0.7\columnwidth,,height = 0.6\columnwidth]{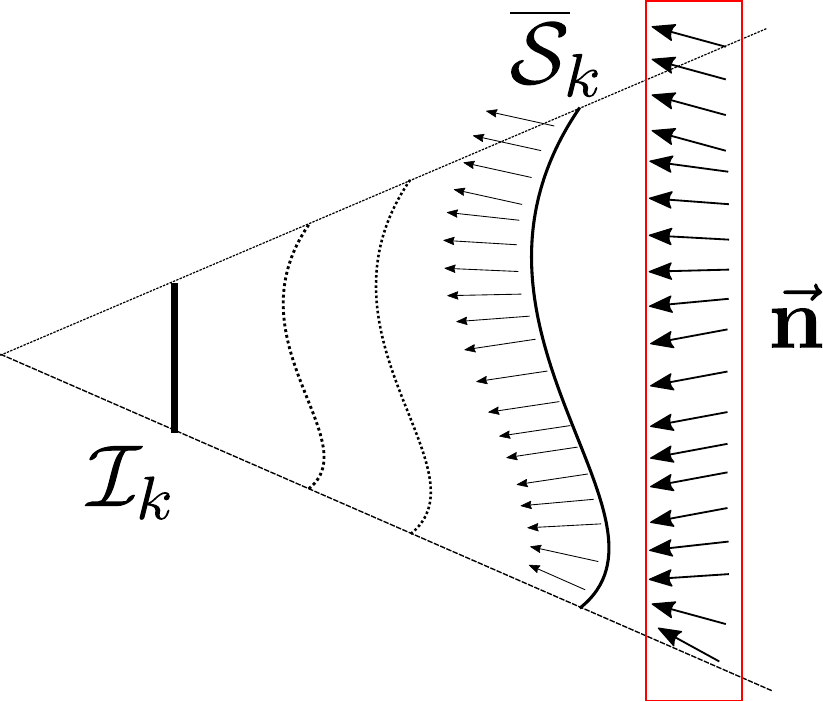}
    \caption{$\overline{\mathcal{S}}_k$ is the estimated up-to-scale surface. $\vec{\mathbf{n}}$ are the set of normals. Two examples of surfaces at a different scale but having the same normals are displayed in dotted lines.  }
    \label{fig:SetofSolutions}
\end{figure}

In contrast to \cite{parashar2017isometric}, optimization (\ref{eq:noptimization}) is incrementally computed. We initialize it with its last estimate achieving a fast convergence. Once the normals are refined in their anchor keyframe, we transfer the normals to the new reference keyframe with eq. \ref{eq:Chris_rel}. We recover the up-to-scale $\overline{\mathcal{S}}_k$ from the set of estimated normals $\vec{\mathbf{n}}$ using Shape-from-Normals (SfN) \cite{chhatkuli2014non}. The surface $\overline{\mathcal{S}}_k$ is regressed with a bicubic b-spline parametrized by its control nodes depth. The control nodes are defined by a regular mesh in the image $\mathcal{I}_k$. We fit the depth of the nodes to obtain a surface orthogonal to the estimated normals with a regularizer in terms of bending energy (Fig.\,\ref{fig:SetofSolutions}). 


\subsection{Surface Alignment}
\label{sec:surfalig}
\begin{figure}
    \centering
    \includegraphics[width=\columnwidth]{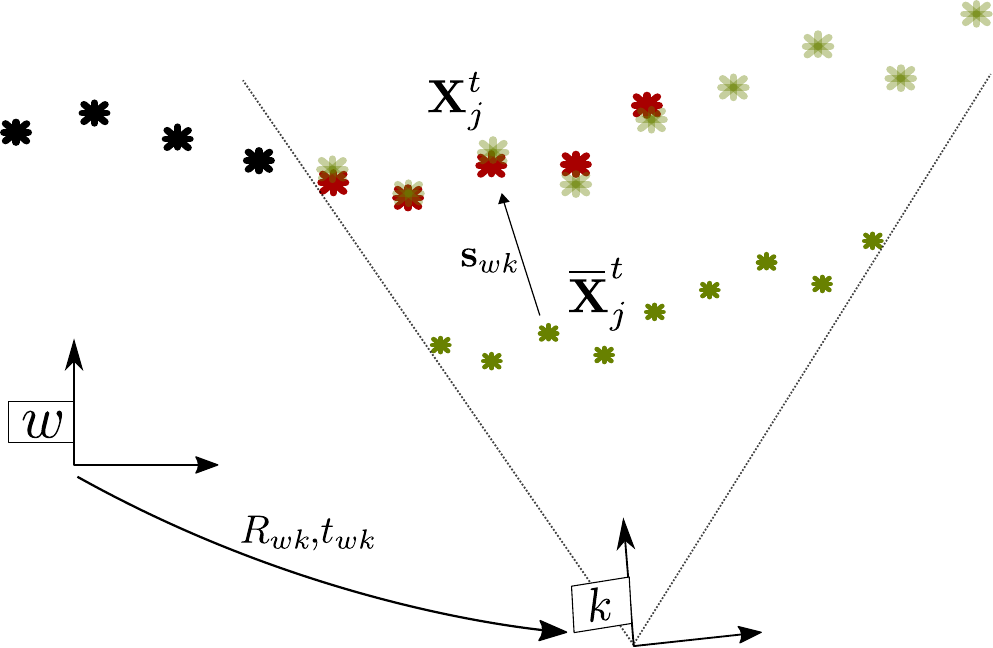}
    \caption{$\mbox{Sim}\left(3\right)$ alignment. We align the the map points $\overline{\mathbf{X}}_j^k \in \overline{\mathcal{S}}_{k} $ of the up-to-scale estimation with the pose of the map points $\mathbf{X}_j^k \in \mathcal{T}_{k-1}^k$ estimated for the frame $k$ deforming the previous template $k-1$}
\label{fig:map_growing}
\end{figure}

The new estimated surface $\overline{\mathcal{S}}_k$ is up-to-scale. We need to recover the solution with a coherent scale ${s}_{wk}$ \textit{wrt}. the already estimated map. This means that the scale-corrected shape-at-rest $\mathcal{S}_{k}$ must have an scale coherent with the deformed template $\mathcal{T}_{k-1}^{k}$ estimated by the tracking when the keyframe was inserted.

We align these surfaces map points through a transformation which belongs the group of similarity of 3-space $\mbox{Sim}\left(3\right)$, by means of  non-linear optimization:
\begin{equation}
    \underset{\mathbf{R}_{wk},\mathbf{t}_{wk},\mathbf{s}_{wk}}{\arg\min} \sum_{j \in \mathbf{X}^k} \left\|\mathbf{s}_{wk} \mathbf{R}_{wk} \overline{\mathbf{X}}_{j}^k  + \mathbf{t}_{wk} - \mathbf{X}_j^k \right\|^2,
\end{equation}
where $\mathbf{R}_{wk}$,$\mathbf{t}_{wk}$,$\mathbf{s}_{wk}$ are the rotation translation and scale defining the $\mbox{Sim}\left(3\right)$ transformation (Fig.\, \ref{fig:map_growing}).

To build our new template $\mathcal{T}_k$, we finally create a triangular mesh from the scale-corrected surface $\mathcal{S}_{k}$ by means of regular triangular mesh in the image. The new map points 3D pose is computed from the matched keypoints by constraining them to be in the estimated surface $\mathcal{S}_{k}$. Then, we embed the re-observed map points and the new map points by projecting them into their corresponding template facet. With this embedding, we calculate the barycentric coordinates of the map points which will be used by the tracking. 

\subsection{Template Substitution}
\label{sec:tempsubs}
Once the surface $S_k$ is computed, the keyframe $k$ is set as the reference keyframe and the current template $\mathcal{T}_{k-1}$ is substituted by $\mathcal{T}_{k}$ computed from $S_k$. The shape observed in the current frame $t$ differs from the shape of the new template $\mathcal{T}_{k}$. This yields to failures in the data association stage, which assumes small deformations, if we substitute the template directly by $\mathcal{T}_{k}$. Therefore, we transfer the matches from $\mathcal{T}_{k-1}^t$ to $\mathcal{T}_{k}$ and compute the current shape $\mathcal{T}_{k}^t$  using optimization \,(\ref{eq:SfTLamarca}).

\subsection{Warp Estimation and Non-Rigid Guided Matching}
\begin{figure}
    \includegraphics[width=\columnwidth]{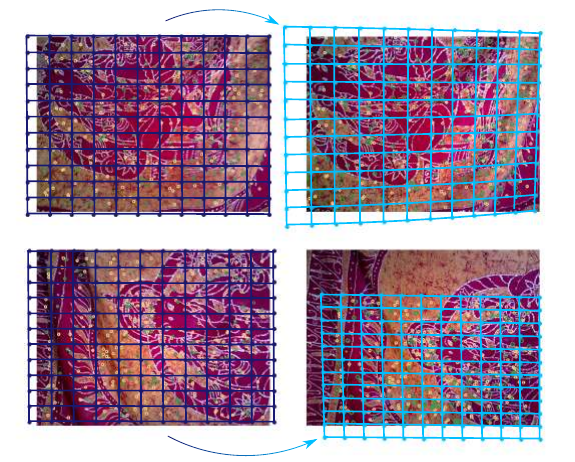}
\caption{Two examples of warp estimation.  Warp estimation between the keyframe $k$ (left) and $k^*$ (right). The warp between $k$ and $k^*$ is plotted in blue. Yellow points are the initially matched map points, green points are the matches added by guided matching stage using the warp.}
\label{fig:keyframe_matching_new_points}
\end{figure}

The input of NRSfM is the set of warps $\eta_{k k^*}$ between an anchor keyframe $k$  and their covisible keyframes $k^*$. The image  warp $\eta_{k k^*}$ is a function that transforms a point in the anchor keyframe into the corresponding point in its covisible $k^*$:

\begin{eqnarray}
\eta_{k k^*}: \left[\hat{x},\hat{y}\right] \in \mathbb{R}^2 &\mapsto& \left[\hat{x}^*,\hat{y}^*\right] \in \mathbb{R}^2 . \nonumber
\end{eqnarray}
We use a particular family of warps called Schwarps \cite{pizarro2016schwarps}, because, as discussed in \cite{parashar2017isometric}, the formulation of the 2D Schwarzian equation regularizers are equivalent to the infinitesimal planarity of the NRSfM. See Figure \,\ref{fig:keyframe_matching_new_points} for two examples of warp between keyframes.

First, we estimate an initial warp between the anchor keyframe $k$ and its covisible keyframe $k^*$ with the matches given by the deformation tracking. Then we use the intial warp to perform a guided matching stage between the keypoints in keyframes $k$ and $k^{*}$.  We accept as a match the keypoint inside a search region with the smallest Hamming distance for the ORB descriptor. We apply a threshold on the ORB similarity to definitively accept a match. Once that we have the new matches we incorporate them to the initial ones and estimate the final warp. See Figure \,\ref{fig:keyframe_matching_new_points} for two examples of warp between keyframes.

\subsection{SLAM Initialization}
\label{sec:init}
At initialization we need to have a template available for the scene surface. We compute it from the first frame of he sequence, assuming its surface $\mathcal{S}_1$, and hence its template $\mathcal{T}_1$ is a plane parallel to perpendicular to the camera optical axis.  

With the second keyframe inserted, the mapping thread starts to compute a new template, that replaces the initial one. The accuracy of the first computed templates strongly depends on how many keyframes are fed in the NRSfM and on how large is the parallax they render.

According to the experiments, our algorithm can track from an inaccurate template with a high quality data association between keyframes, yielding long tracks and a low false positive rate. As a result, as more keyframes rendering high parallax are created, the estimated template eventually converges to the actual scene shape.

\section{Implementation Details}
The method is implemented in C++ and runs entirely on the CPU. We have used the OpenCV library \cite{opencv_library} for base computer vision functions. For the SfT optimization and the LS $\mbox{Sim}\left(3\right)$ registration, we have used the g2o library \cite{kummerle2011g} and its implementation of Levenberg-Marquardt. For the Schwarps optimization, the normal estimation and the shape-from-normals, we have used the Ceres library \cite{ceres-solver}. The runtime depends on the resolution of the mesh used as template. For a mesh of 10$\times$10 nodes the runtime is approximately $50$\,ms for the deformable tracking thread and approximately $400$\,ms for the deformable mapping in a machine with an i7-4700HQ CPU and with 7.7 Gb RAM. The code will is available as a public git repository\footnote{\url{https://github.com/UZ-SLAMLab/DefSLAM}}. 

\section{Experiments}
\label{sec:experiment}
\begin{figure*}[]
  \centering
    \includegraphics[trim = {3cm 0cm 3cm 0cm},clip,width=\textwidth]{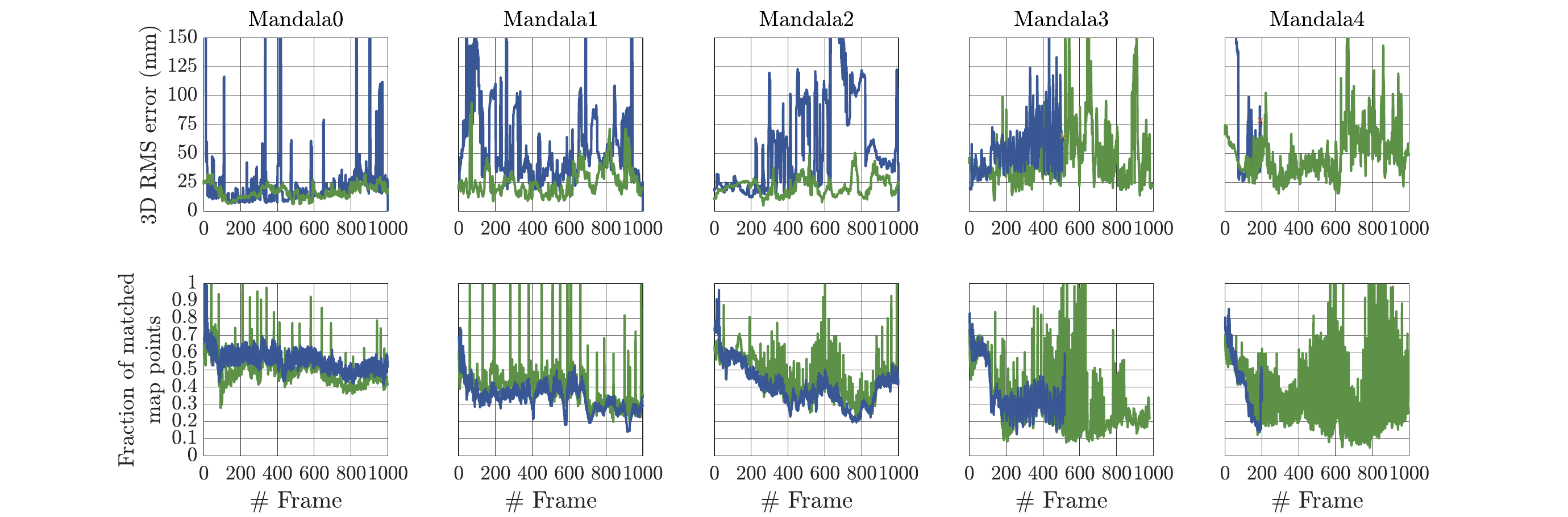}
    \includegraphics[trim = {1cm 0cm 1cm 0cm},clip,width=\textwidth]{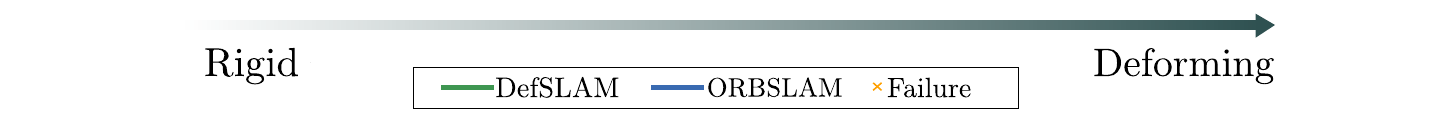}
\caption{
Overall quality for Mandala dataset sequences. From left to right, the scenario contains more deformation. Top: 3D RMS error (mm) per frame (the smaller, the better). Bottom: Fraction of matched map points (the higher, the better).}
\label{fig:deformable_experiment_all_frames}
\end{figure*}
\begin{figure*}[!h]
\centering
    \includegraphics[trim ={1cm 0cm 1cm 3cm},clip,width=\textwidth]{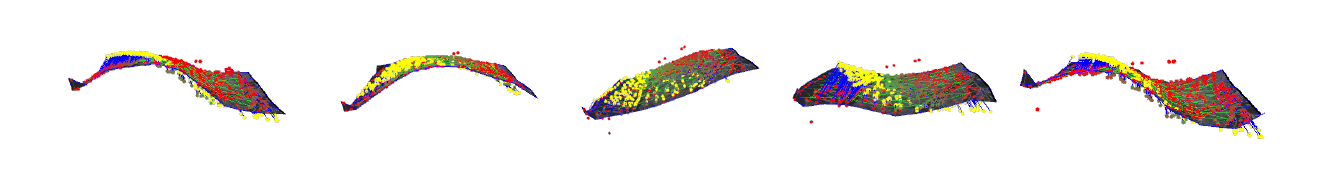}
\caption{Recovering local deformations in the mandala3 sequence. 3D map points in red, 3D point in yellow is the ground truth and blue lines are the difference. DefSLAM can perceive and reconstruct the deforming scene.}
    \label{fig:Mandala}
\end{figure*}
We tested DefSLAM in two datasets. The first dataset is the Mandala dataset which we create to evaluate deformable monocular SLAM  in a laboratory controlled situation. The second is a selection of sequences from the medical Hamlyn dataset (\cite{mountney2010three,stoyanov2005soft}), which comprises a phantom heart, and in-vivo sequences including explartory trajectories. The sequences in both datasets have ground truth depth for each frame, either from stereo or from CT.

We focus on two per frame metrics:  the 3D RMS error of the in-frustum map points and the fraction of  matched map points. The RMS error is computed after a scale alignment for each frame of the sequence, it features the geometrical accuracy. The fraction of  map points matched is the quotient between the map points effectively matched in the current frame, and the number of map points in-frustum of the current frame, i.e maximum number of map points that ideally can be matched. A low fraction signals a poor map that can only represent partially the scene imaged in the current frame.

In addition, we carried out an ablation analysis of the mapping and the tracking. In the mapping, we focused in  NRSfM stages of the normals estimation. In the tracking, we evaluate the performance of the deforming template when compared with a rigid one. We also analyzed the sensitivity of the system to the tuning of the regularizers weights in the tracking optimization (eq. \ref{eq:deformation_Energy}).

Currently, there is no other monocular SLAM for deformable environments to compare with. Thus we select a rigid monocular SLAM method, ORBSLAM \cite{mur2015orb}, as one of the closest for comparison. We had to re-tune several stages of ORBSLAM to process deforming sequences. 1) We relaxed the thresholds for matching and outlier rejection to retain matches despite the deformation. 2) We initialized it with the first frame ground truth map, to avoid the dramatical failure of the monocular intialization. 3) We decreased the rate of new keyframe creation up to one keyframe out of 3 frames, to adapt the map to the scene deformations. On the other hand we compare with MISSLAM \cite{song2018mis} in the Hamlyn phantom heart dataset, as the closest in the medical arena, despite MISSLAM is stereo instead of monocular.

For the sake of repeatability, DefSLAM  was run sequentialized in single-thread, inserting one new keyframe every 10 frames. All the reported results are the median of 5 executions in each sequence.
\subsection{Mandala Dataset}

\begin{figure}
    \centering
    \includegraphics[width = \columnwidth]{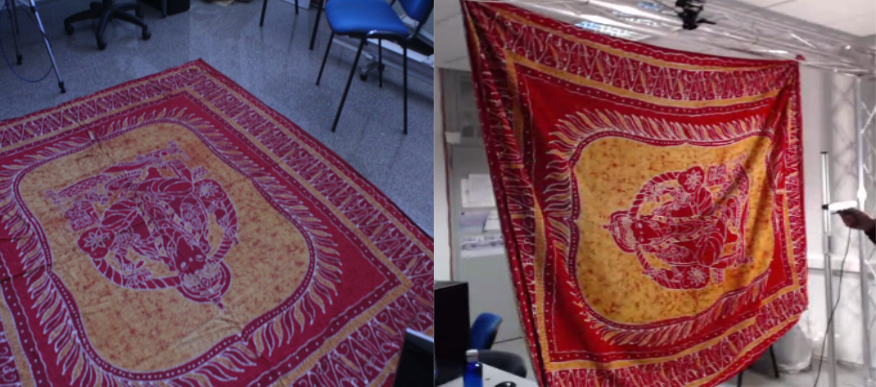}
    \caption{Two configurations of Mandala dataset: rigid planar (Mandala0), and hanged in the rest of the sequences.}
    \label{fig:configurations}
\end{figure}
We introduce the \textbf{Mandala dataset} to evaluate the map quality of deformable monocular SLAM systems in a controlled environment. It is composed of 5 sequences (640x480 pix. at 30 fps) with exploratory trajectories observing a textured kerchief deforming near-isometrically. We increased the hardness of deformation progressively by reducing the period of the waves generated on the kerchief and increasing their amplitude from the shape-at-rest. Fig. \ref{fig:configurations} shows the two configurations: planar and hanged.

In the sequence mandala0, the kerchief remains rigid on the floor. In mandala1, the deformation had an amplitude of $15$\,cm and a period of $2$\,s. In mandala2, the amplitude is $10$\,cm and the period $1$\,s. In the mandala3, the amplitude is $25$\,cm and the kerchief oscillates with a period of  $2$\,s. In the mandala4, the amplitude is $30$\, cm, and its period is halved to $1$\,s.

\subsubsection{Overall quality experiment}
We analyze the overall quality of the estimated map. Figure\,{\ref{fig:deformable_experiment_all_frames}} shows the final results along the five sequences for DefSLAM in green, and ORBSLAM in blue.

In rigid mandala0, DefSLAM obtains a similar 3D RMS error to ORBSLAM. Concerning the fraction of matched map points, both DefSLAM and ORBSLAM got a high percentage which means that the map points that are highly reused, due to the rigidity of the scene. 

In mandala1 and mandala2, the kerchief has low frequency and amplitude deformation. DefSLAM obtains a similar 3D RMS error to the one obtained in mandala0 for both sequences, being able to recover the deformation of the kerchief. ORBSLAM could process the entire sequences, but its 3D RMS error was highly penalized by the deformation, triplicating the error obtained in the mandala0 sequence, and the RMS error of DefSLAM. DefSLAM could recover more accurately the deformation of the scene observed during the sequence both in terms of RMS error and in fraction of matched map points per frame.

In the mandala3 and mandala4 sequences, the conditions are more extreme. ORBSLAM could not process any of these sequences entirely. In this sequences, the fast deformation yields difficulties for DefSLAM which experiments some delay to converge the correct shape. This provoked some peaks in the RMS error. In any case, the error average was around the 4\,cm during both sequences. In Fig.\,{\ref{fig:Mandala}} we can observe the quality of the reconstruction of the local deformations in the sequence mandala3. The fraction of matched map points for DefSLAM was also smaller. Supplementary material includes a video with fragments of the mandala dataset quality results.

\subsubsection{Scale drift analysis}
\label{sec:scaledrift}
\begin{figure}
    \centering
    \includegraphics[trim = {0cm 0cm 0cm 0.25cm},width = 0.75\columnwidth]{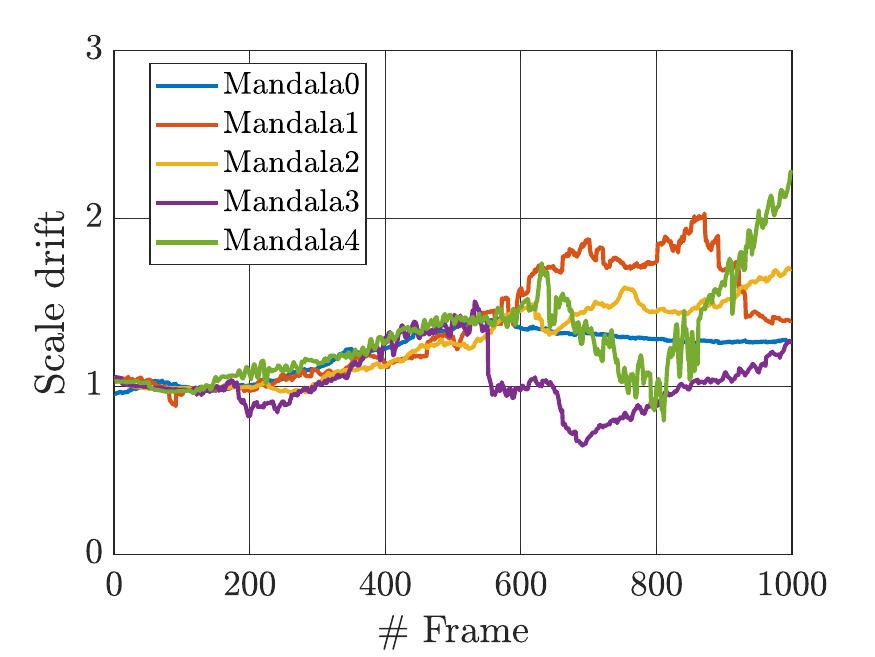}
    \caption{Scale drift along the Mandala sequences. It increases more with more challenging. It is reduced in case of re-observation.}
    \label{fig:Scaledrift}
\end{figure}

The previous section RMSE focuses on the up-to-scale shape accuracy. Fig.\, \ref{fig:Scaledrift} shows the scale drift along the different sequences. The main source of scale drift is the alignment (Sec.\, \ref{sec:surfalig}), where to estimate the scaled template, we align the reference up-to-scale template with the previous reference scaled template. This makes the scale accumulate the misalignment between the new and the old template. The scale drift is close to null in the mandala 0 and increases to higher values to peak the deformation becomes more challenging. Eventually the scale drift can be reduced due to re-observations of the map during the sequence.

\subsubsection{Sensitivity Analysis}
\label{sec:rigid}
\begin{figure}
    \centering
    \includegraphics[trim = {0cm 0cm 0cm 0.75cm},width = 0.75\columnwidth]{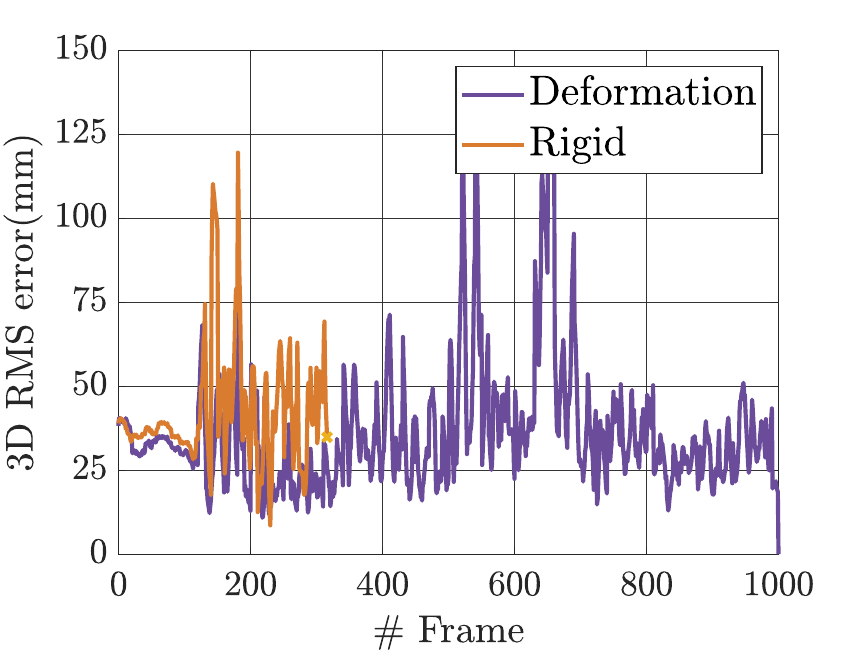}
    \caption{Rigid tracking vs deformation tracking surface error as 3D RMS scene reconstruction error per frame in mm.}
    \label{fig:deftrackvsrigid}
\end{figure}
All the experiments reported, both in the Mandala dataset and Hamlyn, were run with $\lambda_s =16000$, $\lambda_b = 300$ and $\lambda_r=0.02$ as standard tuning.

To better understand the role of the weights, we varied their values to study their effect in the final 3D RMS error and scale drift in the challenging mandala3. We run the entire sequence and evaluated the RMS error at the end of the sequence from frames \# 800 to \# 1000. The error is not servery affected, remaining between 20 and 40 m, for a range of values from $\lambda_s = [1600,100000]$, $\lambda_b = [100,1000]$, $\lambda_r=[0,0.1]$. By decreasing the $\lambda_s$ and $\lambda_b$ values, the system becomes unconstrained and fails in process the entire sequence. By increasing $\lambda_s$ and $\lambda_b$, the system assumes rigidity thus causing another failing scenario. Figure \,\ref{fig:deftrackvsrigid} shows the extreme case of a perfectly rigid and fixed template compared with our standard tuning. It can be seen how a rigid template for tracking fails to survive strong scene deformations. This case correspond to high values for the three coeficients $\lambda_s$, $\lambda_b$ and $\lambda_r$. 

The reference regularizer has proven critical to reduce the scale drift specially in the Hamlyn SeqHeart sequence where the camera is imaging constantly the same zone and observing the entire template with few boundary point constraints (Sec. \ref{sec:Hamlyn}), from  36\,\%  for $\lambda_r$ to 2\,\% for $\lambda_r=0.02$

\subsubsection{Deformation mapping normal estimation accuracy}
\begin{figure}[]
    \centering
    \begin{subfigure}
    \centering
    \includegraphics[trim ={2cm 0cm 2cm 0.75cm},clip,width=\columnwidth]{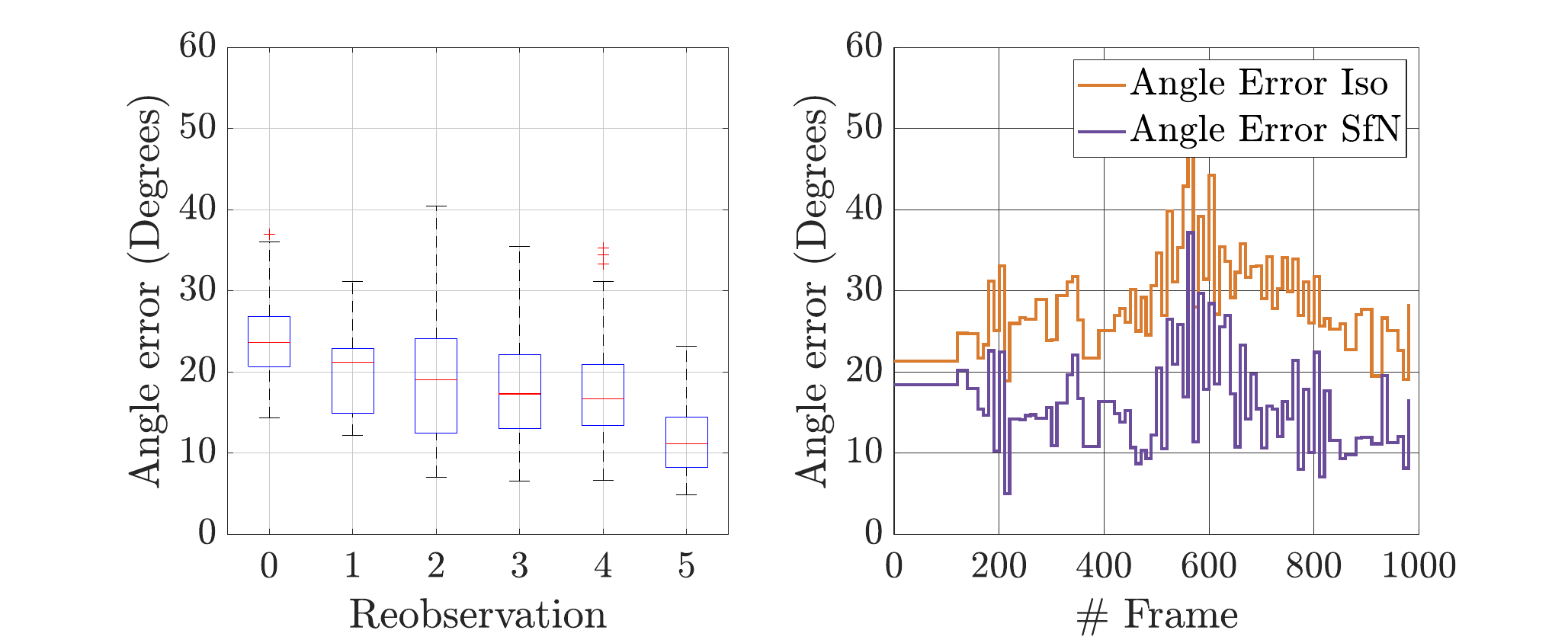}
    \end{subfigure}
    \caption{(Left) Box-and-whisker plot for the normals angle error in a keyframe after SfN, improvement as a function of the keyframe resobservations. (Right) per keyframe RMSE angle error for the normal orientation after NRSfM and after SfN.}
    \label{fig:KeyFrame_error}
\end{figure}

We analyze the quality of the deformation mapping for sequence mandala3 focusing in the angle error between the estimated normal and the ground truth normal, in the two stages of the normal estimation, the initial NRSfM and the subsequent SfN (Sec.\,\ref{sec:incremOpt}). 
\begin{figure*}[h]
  \centering
    \includegraphics[trim = {3cm 0cm 3cm 0cm},clip,width=\textwidth]{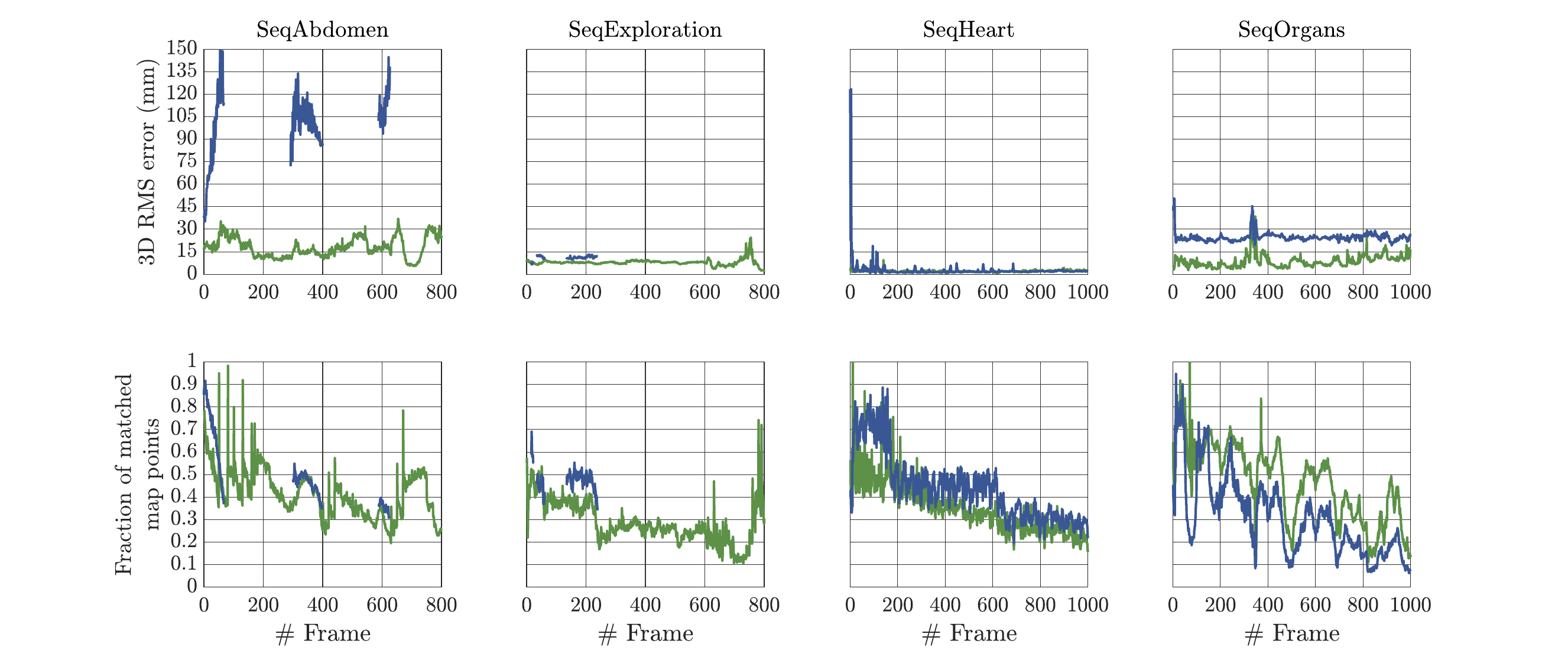}
      \includegraphics[trim = {1cm 7cm 1cm 1.9cm},clip,width=\textwidth]{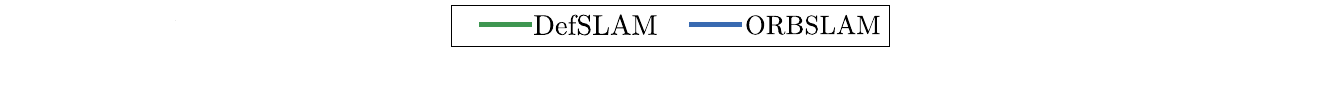}
    \caption{Processing Hamlyn sequences. Green DefSLAM, blue ORBSLAM. From left to right: Heart, organs, abdomen and exploration sequences. Per frame RMS scene reconstruction error in mm after a per frame scale alignment with the stereo ground truth.}
\label{fig:medical_experiment}
\end{figure*}
\begin{figure*}
 \includegraphics[width=\textwidth]{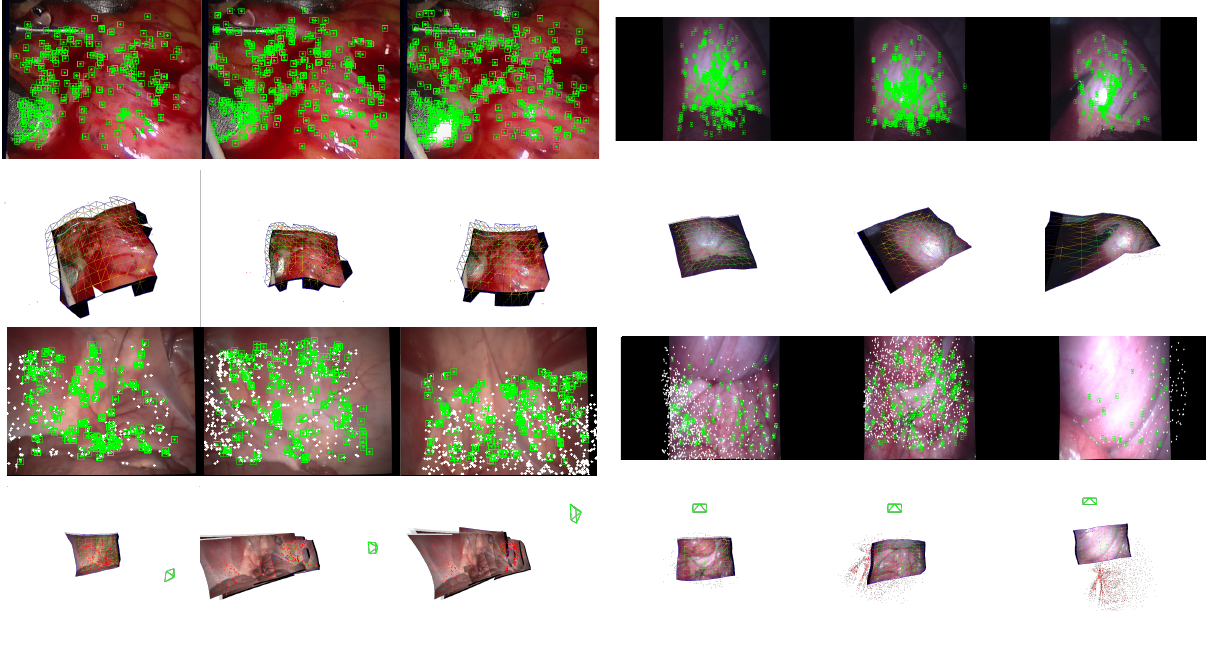}
    \caption{DefSLAM in in-vivo Hamlyn dataset sequences. 3 typical 2D images an the corresponding 3D maps. (Top left) Heart sequence. (Top right) Organs Sequence. (Bottom left) Abdominal sequence. (Bottom right) Exploration sequence.}
    \label{fig:HamlynImages}
\end{figure*}
\begin{figure}
    \centering
    \includegraphics[trim = {0cm 0cm 0cm 0.25cm},width = 0.75\columnwidth]{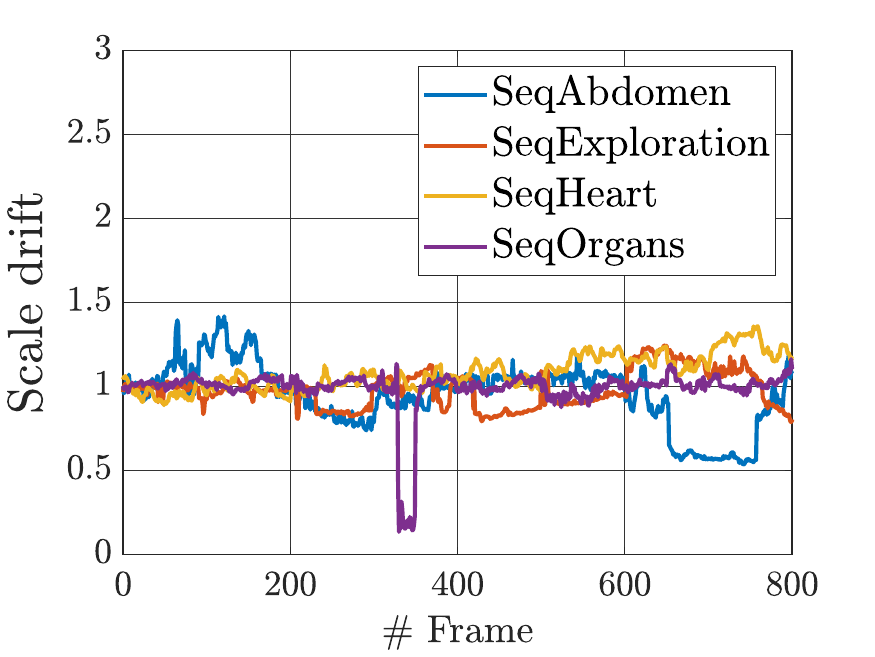}
    \caption{Scale drift along the Hamlyn dataset sequences.}
    \label{fig:ScaledriftMedical}
\end{figure}
Figure\,{\ref{fig:KeyFrame_error}} shows the RMS angle error of the shape estimated by the NRSfM versus the error after the SfN stage. SfN consistently reduces the error through the entire sequence improving the normals. Averaging the error for all the keyframes in the sequence the SfN achieves a 15 deg RMSE versus the 22 deg of the NRSfM.  

The output of the NRSfM is the set of surface normals for each map point in the reference keyframe. The normal of a map point is reestimated after each re-observation of that point in a new keyframe. Figure \,\ref{fig:KeyFrame_error} shows the evolution of the RMS angle error for the normals in a keyframe along 5 re-observations after its creation. We can see how the median error goes from 23 degrees at initialization down to 12 deg after the 5th re-observation.

\subsection{Hamlyn Dataset}
\label{sec:Hamlyn}

Our last experiments test DefSLAM in intracorporeal in six sequences from the \emph{Hamlyn dataset} \cite{mountney2010three,stoyanov2005soft} to evaluate our algorithm in medical images. The first two sequences are recorded with a ex-vivo phantom heart \cite{stoyanov2010real} syncronised with a CT scanner to register ground truth. In addition, we processed four in-vivo laparoscopic sequences (See Fig. \ref{fig:HamlynImages}): 1) SeqAbdomen is an exploration of the abdominal wall where the scene remains almost rigid (Fig.\,\ref{fig:HamlynImages} Bottom left). 2) SeqExploration performs an exploration around the exterior of the bowel with low texture. it has a small deformation at the beginning (Fig.\,\ref{fig:HamlynImages} Bottom right). 3) SeqHeart \cite{stoyanov2005soft} is a non-rigid beating heart observed by a fixed camera. 4) SeqOrgans is an abdominal exploration and deformation of the scene due to tool interfering (Fig.\,\ref{fig:HamlynImages} Top right). 

The closest SLAM system to ours reporting accuracy \textit{wrt.} an external sensor in medical sequences is MISSLAM \cite{song2018mis}. We evaluate our system in the same sequences, i.e. the ex-vivo phantom heart sequences. Despite the  lack of camera motion, the scenes have enough deformation for DefSLAM to reconstruct them. We report a mean accuracy of 3 and 4\,mm in the sequence phantom5 and phantom7, respectively. 
The average accuracy MISSLAM as reported by the authors in \cite{song2018mis} is 0.28\,mm and 0.35\,mm. Concerning the execution time, we report a similar runtime per frame, but DefSLAM runs in CPU unlike MISSLAM that uses GPU. It has to be noted that they use stereo input in contrast with DefSLAM which is a purely monocular method.

Fig.\,\ref{fig:medical_experiment} reports the median of 5 executions RMS error during the four in-vivo Hamlyn sequences and Fig.\,\ref{fig:ScaledriftMedical} shows its corresponding scale drift. As it happened with the Mandala dataset (Sec.\,\ref{sec:scaledrift}, the scale drift got slighly increased for the more challenging sequence. In the sequence where the camera in the same zone there is no scale drift.

DefSLAM is able to process SeqAbdomen and SeqExploration entirely with a mean 3D RMS error of 17 mm and 10 mm respectively. In these scenes, the camera explore but it come back to the same zone. DefSLAM was able to re-observe part if the map already built and thus reduced the scale drift. ORBSLAM performed poorly in this sequences and could not process them entirely. 

In SeqHeart, the camera is practically static, but DefSLAM was able to initialize with the monocular strategy proposed even with a short parallax. The 3D RMS error was  approximately 3\,mm, equal to the ex-vivo phantom result with a much better groundtruth. ORBSLAM initializated with the ground truth was able to process the entire sequence with an error of 5\,mm

Finally in the sequence SeqOrgans, DefSLAM shows its ability to perform the reconstruction of a deformable scene in exploratory sequence with an accuracy of 8\,mm. It survives to the tool clutter that cover almost entirely the image, correcting the scale drift. In the end of the sequence, the tool deforms the organ imaged and DefSLAM was able to recover the deformation of the scene with the same error than in the rest of the sequence.

Fig.\,\ref{fig:HamlynImages} shows the overall quality of the 3D reconstruction of the medical sequences. Supplementary material includes the video with the results in all the sequences.

\section{Conclusions and Future Work}
We have formulated DefSLAM, the first deformable SLAM able to process monocular sequences. We have proposed to split the computation of DefSLAM in two parallel threads. The deformation tracking thread is devoted to estimating the camera pose and the deformation of the scene, it is based on SfT. SfT needs a prior of the geometry of the scene encoded in the template. When exploring new zones, our method estimates new templates to cover new areas. Our second thread, the deformation mapping, is devoted to periodically re-estimating the template to better adapt it to the currently observed scene. Both SfT and NRSfM model the cameras as perspective, hence the system is able to handle close-ups typical in scene exploration where perspective effects are prevalent. 

Our experiments confirm that the proposed method is able to handle real exploratory trajectories of a deforming scene. Direct comparison with other systems is not possible, we have focused the comparison with the rigid monocular ORBSLAM after its re-tuning to handle non-rigid scenes. This comparison proves that  DefSLAM is able to robustly initialize from monocular sequences,  continuously adapt the map to the scene deformation, and producing  accurate scene estimates. 

We have also shown in preliminary experiments that the system is able to handle medical endoscopy images. The next step will be its adaptation for medical imagery to handle all kinds of challenges not taken into account in the present work, \textit{i.e.} uneven illumination, poor visual texture, non-isometric deformations or ultra close-up shots exploring the endoluminal cavities. 

Another future work is to develop a full fledged mapping system including multiple maps,  relocalization, loop closure or long term place recognition to achieve robust performance for extended periods of time or multiple moving and deforming bodies.
\section*{Acknowledgement}
This work was supported the European Union’s Horizon 2020 research and innovation
programme under grant agreement No 863146, the  PGC2018-096367-B-I0 MCIU/AEI/FEDER, UE, the Spanish Agencia estatal de investigación DPI2017-91104-EXP and the MINECO scholarship BES-2016-078678.
{\small
\bibliographystyle{ieee}
\bibliography{main}
}
\appendix
\subsection{Derivatives of Regularizers}
\label{FirstAppendix}
We show the Jacobian terms of the regularizers to prove that it does not have singularities:
\paragraph{Stretching}
The streching error $e_{s}(\mathcal{L}_{t}^k,\mathcal{T}_{k})_e$ for the edge $e$ is: 
\begin{equation}
 e_{s}(\mathcal{L}_{t}^k,\mathcal{T}_{k})_e = \left(\frac{l_e^t-l_e^k}{l_e^k}\right),    
\end{equation}
being
\begin{equation}
l_e^t = \|(\bm{V}_{e^1}^t-\bm{V}_{e^2}^t)\|_2
\end{equation}
where $\bm{V}_{e^1}^t$ and $\bm{V}_{e^2}^t$ are the two nodes of the edge e in the instant $t$.
Its derivative is
\begin{equation}
  \frac{\partial e_{s}(\mathcal{L}_{t}^k,\mathcal{T}_{k})_e}{\partial \bm{V}_{e^i}^t} = \frac{(\bm{V}_{e^1}^t-\bm{V}_{e^2}^t)}{l_e^k l_e^t}  
\end{equation}
\paragraph{Bending}
The bending error $e_{b}(\mathcal{L}_{k}^t,\mathcal{T}_{k})_n$ for the node $n$ connected with its neighbours $\bm{V}_l \in \mathcal{N}_j$ through the edge $e_l$ is: 

\begin{equation}
       e_{b}(\mathcal{L}_{k}^t,\mathcal{T}_{k})_n = \frac{\delta_{n}^{t}-\delta_{n}^{k}}{l_{e_l}^{k}}.
\end{equation}
where $\bold{\delta}_{n}^{t}$ is the mean curvature of the surface at the instant t. It is estimated though the neighbours of the node and itself. 
\begin{equation}
\label{LaplacianCoords}
\bm{\delta}_{n}^{t} = \bm{V}_{n}^{t} - \frac{1}{\sum_{l \in \mathcal{N}_j} \omega_l}\sum_{l \in \mathcal{N}_j} \omega_l \bm{V}_{l}^{i} 
\end{equation}
\begin{equation}
\delta_{n}^{t} = \|\bm{\delta}_{n}^{t} \|_2
\end{equation}
We assume fixed the values of the weights.
Its derivative respect the node $\bm{V}_{n}^t$ is:
\begin{equation}
    \frac{\partial e_{b}(\mathcal{L}_{k}^t,\mathcal{T}_{k})_n}{\partial \bm{V}_{n}^t} = \frac{\bm{\delta}_{n}^{t}}{l_{e_l}^{k} \delta_{n}^{t}} 
\end{equation}
with respect to its neighbours $\bm{V}_{l}^t$
\begin{equation}
    \frac{\partial e_{b}(\mathcal{L}_{k}^t,\mathcal{T}_{k})_n}{\partial \bm{V}_{l}^t} = \frac{\omega_l}{\sum_{l \in \mathcal{N}_j} \omega_l} \frac{\bm{\delta}_{t}^{n}}{l_{e_l}^{k} \delta_{t}^{n}} 
\end{equation}
In case of being a plane, the mean curvature and its derivative tends to zero.
\begin{equation}
     \frac{\partial e_{b}(\mathcal{L}_{k}^t,\mathcal{T}_{k})_n }{\partial \bm{V}_{n}^t} = 0        \,, \quad     \bm{\delta}_n^t = 0 
\end{equation}

\paragraph{Reference}
The reference error is
\begin{equation}
e_{r}(\mathcal{L}_{k}^t,\mathcal{L}_{k}^{k}) = \mathbf{V}_{n}^{t} -\mathbf{V}_{n}^{k}.
\end{equation}
and its derivative:
\begin{equation}
   \frac{\partial e_{r}(\mathcal{L}_{k}^t,\mathcal{T}_{k}) }{\partial \bm{V}_{n}^t}  = 1.
\end{equation}

\begin{IEEEbiography}[{\includegraphics[width=1in,height=1.25in,clip,keepaspectratio]{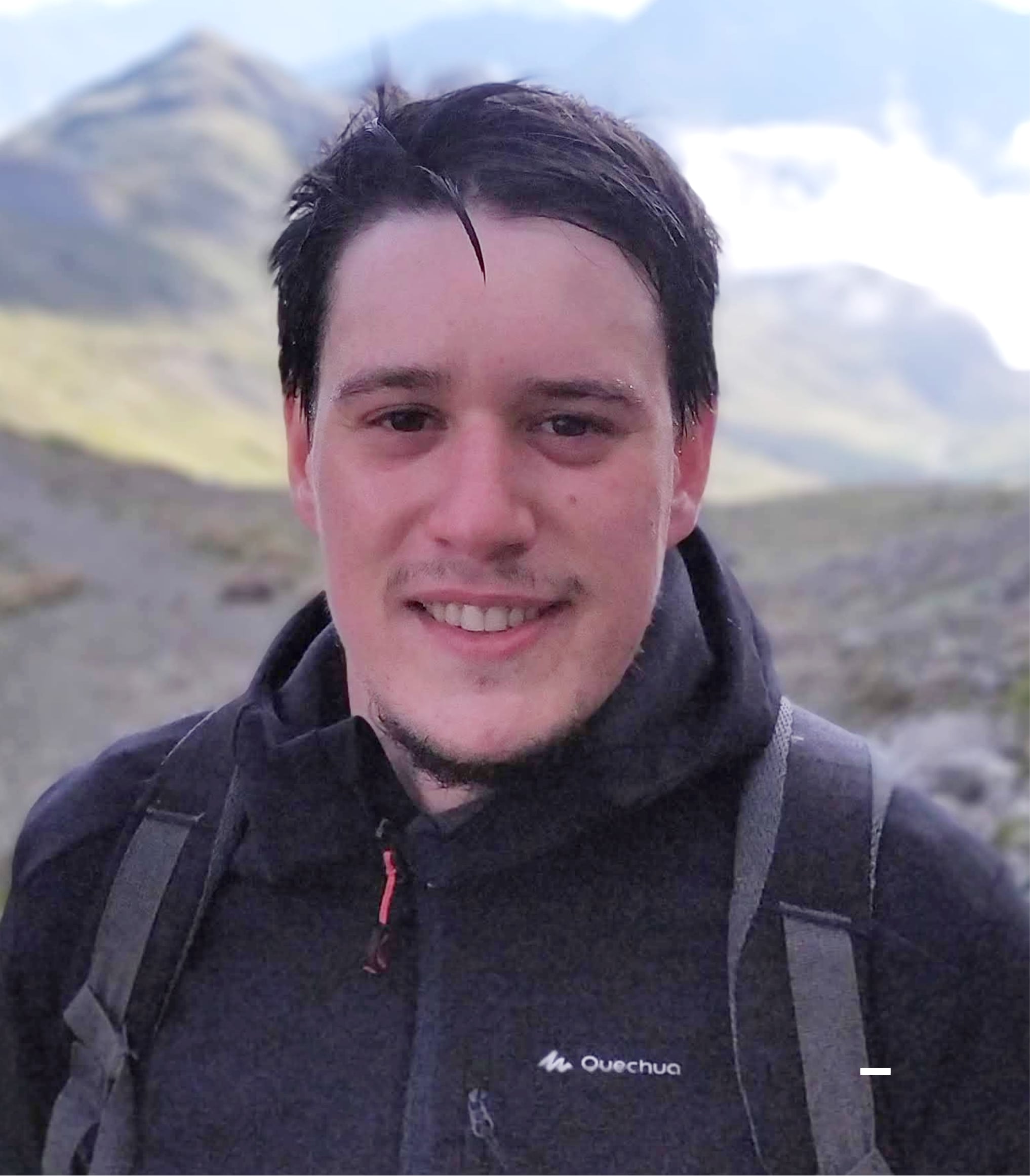}}]%
{Jose Lamarca} received a Bachelor's and M.S. Degree in Industrial Engineering (mention in Robotics and Computer Vision) from Universidad de Zaragoza, where he is currently PhD. student in the I3A Robotics, Perception and Real-Time Group. His research interests are real-time visual SLAM for rigid and deformable environments.
\end{IEEEbiography}

\begin{IEEEbiography}[{\includegraphics[width=1in,height=1.25in,clip,keepaspectratio]{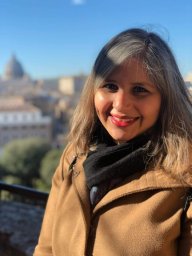}}]%
{Shaifali Parashar}
received her Ph.D. in Computer Vision from the Universite d’Auvergne in 2017. She is currently a PostDoc researcher at CVLab, EPFL. Her research interests are 3D computer vision including non-rigid 3D reconstruction and deformable SLAM.
\end{IEEEbiography}

\begin{IEEEbiography}[{\includegraphics[width=1in,height=1.25in,clip,keepaspectratio]{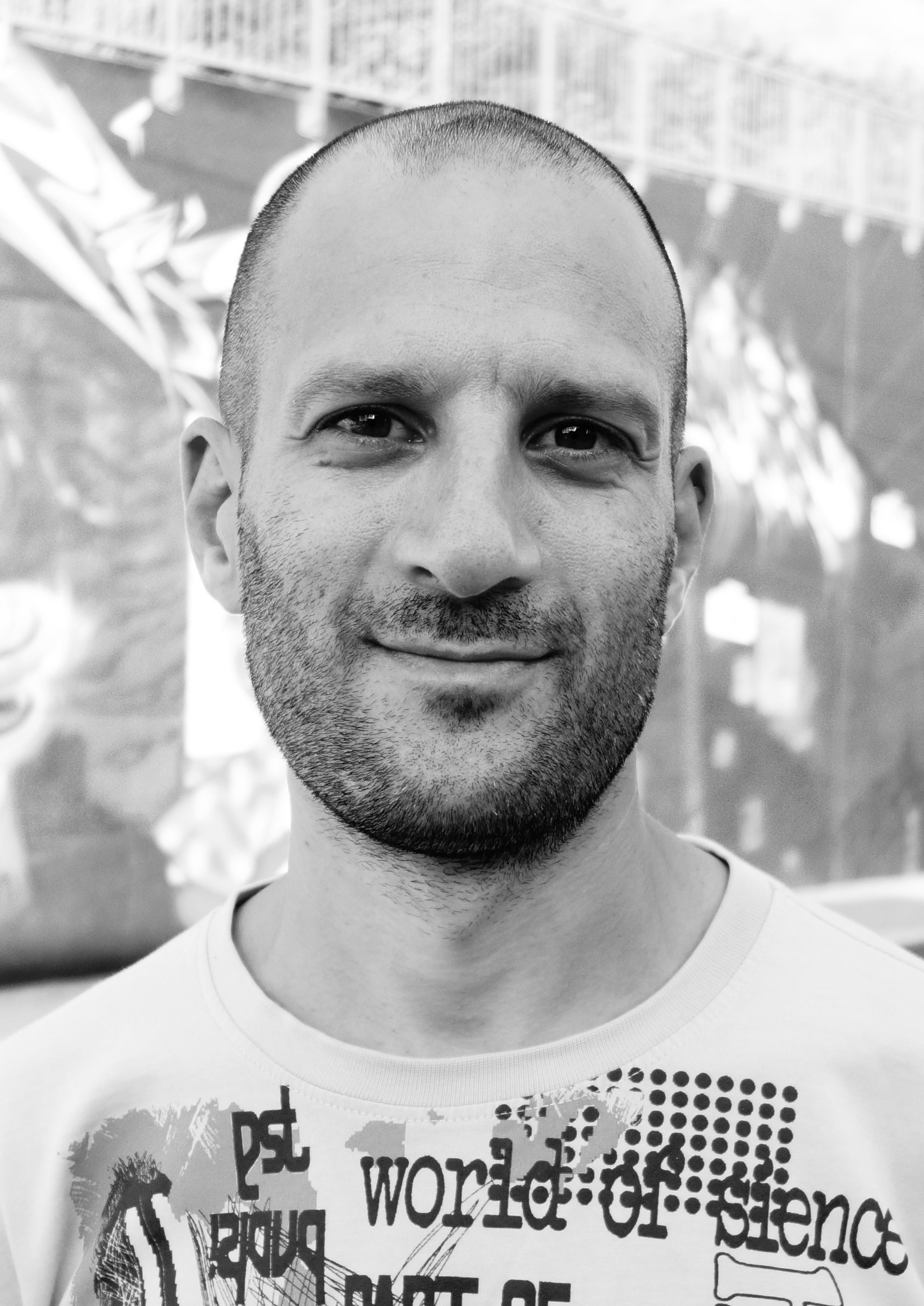}}]%
{Adrien Bartoli}
has held the position of Professor of Computer Science at Université Clermont Auvergne since fall 2009 and has been a member of Institut Universitaire de France (2016-2021). He leads the Endoscopy and Computer Vision (EnCoV) research group at the University and Hospital of Clermont-Ferrand. His main research interests are in computer vision, including image registration and Shape-from-X for rigid and deformable environments, and their application to computer-aided medical interventions.
\end{IEEEbiography}

\begin{IEEEbiography}[{\includegraphics[width=1in,height=1.25in,clip,keepaspectratio]{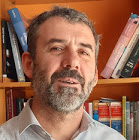}}]%
{J. M. Martínez Montiel} (Arnedo, Spain, 1967). He received the M.S. and PhD degrees in electrical engineering from Universidad de Zaragoza, Spain, in 1992 and 1996, respectively. He has been awarded several Spanish MEC grants to fund research with the University of Oxford, U.K., and with Imperial College London, U.K.

He is currently a Full Professor with the Departamento de Informática e Ingeniería de Sistemas, Universidad de Zaragoza, where he is in charge of perception and computer vision research grants and courses. His interests include real-time visual SLAM for rigid and non-rigid environments, and the transference of this technology to robotic and non-robotic application domains. He has received several awards including the IEEE Transactions on Robotics King-Sun Fu Memorial Best Paper Award 2016. Since 2020 he is coordinating the EU FET EndoMapper grant aimed to bring visual SLAM to medical intracorporeal scenes.
\end{IEEEbiography}

\end{document}